\documentclass[12pt,letterpaper]{article}
\usepackage[a4paper, total={7in, 10in}]{geometry}

\usepackage{helvet}
\usepackage{authblk}
\usepackage{hyperref}
\usepackage{lineno}
\usepackage{graphicx}

\usepackage{amsmath}
\usepackage{multirow}
\usepackage{amsfonts}
\usepackage{booktabs}

\usepackage{makecell}
\usepackage{array}
\usepackage{enumitem}

\usepackage{algorithm}
\usepackage{algorithmic}

\usepackage{subfigure}

\usepackage{smartdiagram}
\usepackage{tikz}
\usetikzlibrary{shapes,arrows}
\usepackage[edges]{forest}
\usepackage{url}

\makeatletter
\def\@maketitle{
  \newpage
  \null
  \vskip 2em%
  \begin{center}%
  \let \footnote \thanks
    {\Large \bfseries \@title \par}%
    \vskip 1.5em%
    {\normalsize
      \lineskip .5em%
      \begin{tabular}[t]{c}%
        \@author
      \end{tabular}\par}%
    \vskip 1em%
    {\normalsize \@date}%
  \end{center}%
  \par
  \vskip 1.5em}
\makeatother

\title{From Challenges and Pitfalls to Recommendations and Opportunities: Implementing Federated Learning in Healthcare}
\date{}

\author[1,2]{Ming Li}
\author[3,4]{Pengcheng Xu}
\author[2]{Junjie Hu}
\author[1,5]{Zeyu Tang}
\author[1,2,6,7,*]{Guang Yang}

\affil[1]{Bioengineering Department and Imperial-X, Imperial College London, London W12 7SL, UK}
\affil[2]{National Heart and Lung Institute, Imperial College London, London SW7 2AZ, UK}
\affil[3]{Athinoula A. Martinos Center for Biomedical Imaging, Massachusetts General Hospital, Boston, Massachusetts, USA}
\affil[4]{State Key Laboratory of Extreme Photonics and Instrumentation, College of Optical Science and Engineering, Zhejiang University, Hangzhou, China}
\affil[5]{Tri-Institutional Computational Biology \& Medicine Program, Weill Cornell Medicine of Cornell University, New York, USA}
\affil[6]{Cardiovascular Research Centre, Royal Brompton Hospital, London SW3 6NP, UK}
\affil[7]{School of Biomedical Engineering \& Imaging Sciences, King's College London, London WC2R 2LS, UK}
\affil[ ]{\texttt{ming.li@imperial.ac.uk, pengchengxu@zju.edu.cn, j.hu@imperial.ac.uk, zet4004@med.cornell.edu}}
\affil[*]{Correspondence: \texttt{g.yang@imperial.ac.uk}}

\usepackage[super,comma,sort&compress]{natbib}

\begin{document}

\maketitle
\sloppy

\section*{Abstract}
Federated learning holds great potential for enabling large-scale healthcare research and collaboration across multiple centres while ensuring data privacy and security are not compromised.
Although numerous recent studies suggest or utilize federated learning based methods in healthcare, it remains unclear which ones have potential clinical utility.
This review paper considers and analyzes the most recent studies up to May 2024 that describe federated learning based methods in healthcare.
After a thorough review, we find that the vast majority are not appropriate for clinical use due to their methodological flaws and/or underlying biases which include but are not limited to privacy concerns, generalization issues, and communication costs.
As a result, the effectiveness of federated learning in healthcare is significantly compromised.
To overcome these challenges, we provide recommendations and promising opportunities that might be implemented to resolve these problems and improve the quality of model development in federated learning with healthcare.
\section*{Keywords}
Federated Learning, Healthcare, Pitfalls, Challenges, Recommendations, Opportunities.

\section{Introduction}
\label{Introduction}
The integration of Artificial Intelligence (AI) into healthcare research has started a transformative era, catalyzing unprecedented advancements in patient care, diagnostic precision, and therapeutic efficacy~\citep{liu2021deep}. 
However, developing robust AI models requires a vast amount of multi-centre data.
A notable example is the genome-wide association studies, when confined to data from a single institution, are often limited by sample size, failing to identify established biomarkers~\citep{newton2013validation}. 
This underscores the imperative for collaborative data sharing among institutions.
Standard AI approaches rely on centralized datasets for model training, but in healthcare, centralization is complex due to various factors such as privacy concerns, regulatory constraints, as well as legal, ethical and technological barriers to data sharing~\citep{ngiam2019big}.
\par

Federated Learning (FL) emerges as a revolutionary paradigm, promising the collaborative training of AI models across distributed datasets without data sharing~\citep{mcmahan2017communication}.
By enabling privacy-preserving data analysis across multiple data silos, FL can exploit the full potential of worldwide healthcare data across different demographics, unlocking insights unattainable by isolated institutions.
Models trained in a federated fashion are potentially able to yield even less biased decisions and higher sensitivity to rare cases as they are exposed to a more complete data distribution.
Recent studies have shown that models trained by FL can achieve performance comparable to the ones trained on centrally hosted datasets and superior to models that only see isolated single-institutional data~\citep{sheller2019multi,sheller2020federated}.
Notably, early studies into FL, particularly in areas like brain tumour~\citep{sheller2020federated}, triple negative breast cancer~\citep{ogier2023federated} and COVID-19~\citep{dayan2021federated}, also have begun to illustrate the potential for generalizability beyond a single institution.
\par

Today's pioneering large-scale initiatives span academic research, clinical applications, and industrial translations, collectively advancing FL in healthcare.
\textit{Within academic research,} consortia such as Trustworthy Federated Data Analytics (TFDA)~\citep{tfda2024} and the~\cite{jip2024} spearhead decentralized research across institutions. An illustrative example is the international collaboration employing FL to develop AI models for mammogram assessment, which outperformed single-institutional models and exhibited enhanced generalizability~\citep{nvidia2020}.
\textit{Moving to clinical applications,} projects like HealthChain~\citep{healthchain} and DRAGON~\citep{dragon} aim to deploy FL across multiple hospitals in Europe, facilitating the prediction of treatment responses for cancer and COVID-19. By aiding clinicians in treatment decisions based on histology slides and CT images, FL demonstrated direct clinical impact. Another large scale project is the Federated Tumour Segmentation (FeTS) initiative~\citep{fets}, which involves 30 institutions globally, that utilize FL to improve tumour boundary detection across various cancers.
\textit{In the industrial domain,} collaborative efforts like~\cite{melody} demonstrate how competing companies can optimize the drug discovery process through multi-task FL while protecting their proprietary data.
\par

Despite FL's promising advantages, integrating it within healthcare still faces methodological flaws and underlying biases.
These encompass but are not limited to, addressing privacy concerns~\citep{dayan2021federated,fang2022dp}, generalization issues~\citep{kline2022multimodal}, communication costs~\citep{rothchild2020fetchsgd}, and the non-independent and identically distributed (non-IID) nature of healthcare data across institutions~\citep{nguyen2021federated}, safeguarding patient data against sophisticated inference attacks that could potentially deanonymize sensitive information from model updates~\citep{bouacida2021vulnerabilities}, and the necessity for standardization across FL implementations. 
Moreover, there's a pressing need for models that not only exhibit robust performance across diverse datasets but are also interpretable and transparent in their predictions and decision-making processes~\citep{li2023towards,li2022explainable}.
\par

To facilitate the implementation of FL in healthcare, we have considered and analyzed the most recent studies, delving into the practical application of FL in healthcare.
We provide numerous recommendations and promising opportunities, which, if followed appropriately, might be able to mitigate current pitfalls and challenges, ultimately leading to high-quality development and reliable reporting of results in FL with healthcare.
Our review makes contributions as follows:
\begin{itemize}
    \item Quantifying and evaluating the integrity and variation of most recent and advanced FL technologies in healthcare to identify challenges, flaws and pitfalls;
    \item Providing a taxonomized, in-depth analysis and discussion of various aspects of FL within healthcare;
    \item Offering evidence-based guidelines and recommendations to enhance the quality of FL development, ensuring fair and reproducible comparisons of FL strategies, while also identifying emerging trends and suggesting future opportunities for improving patient outcomes and streamlining clinical workflows.
\end{itemize}
\par

The rest of this review is structured as follows. 
\textit{Section 2} provides an overview of the background and preliminaries of FL. 
\textit{Section 3} describes the screening procedure adopted in this work. 
\textit{Section 4} highlights the key findings of our analysis. 
\textit{Section 5} explores recent advances, challenges, and pitfalls in implementing FL in healthcare, offering practical recommendations to overcome current limitations and outlining potential future research directions.
\par

\begin{center}
\begin{minipage}{0.8\linewidth}
\begin{algorithm}[H]\small
    \caption{\small FL with FedAvg Training process. $K$ clients are indexed by $k$; $C$-fractions of clients are selected at each round; $E$ is the number of local epochs; $B$ is the local mini-batch size; $\eta$ is the learning rate.} 
    \hspace*{0.013in} {\bf \textit{ServerExecutes:}} 
    \begin{algorithmic}
        \STATE initialize the parameters of model $\theta_{0}$
        \FOR{each round $t=1,2,...,T$} 
        \STATE $m \leftarrow$max$(C\cdot K, 1)$
        \STATE $S_{t} \leftarrow$ (random set of $m$ clients)
            \FOR{each client $k \in S_{t}$ \textbf{in parallel}}
            \STATE $\theta_{t+1}^{k} \leftarrow$ ClientUpdate$(k,\theta_{t})$
            \ENDFOR
        \STATE $\theta_{t+1} \leftarrow \sum_{k=1}^{K} w_k \theta_{t+1}^{k}$
        \ENDFOR
    \end{algorithmic}
    \vfill
    \hspace*{0.05in} {\bf \textit{ClientUpdate$(k,\theta)$:}}
    run on client $k$
    \begin{algorithmic}
        \STATE $\mathcal{B} \leftarrow$ (split $\mathcal{P}_{k}$ into batches of size $B$)
        \FOR{each local epoch $i$ from $1$ to $E$} 
            \FOR{batch $b \in \mathcal{B}$}
            \STATE $\theta \leftarrow \theta - \eta \nabla l(\theta; b)$
            \ENDFOR
        \ENDFOR
        \RETURN $\theta$ to server
    \end{algorithmic}
    \label{algorithm_fl_basic}
\end{algorithm}

\end{minipage}
\end{center}

\section{Preliminaries}
FL, introduced in 2017~\citep{mcmahan2017communication} as federated averaging (FedAvg), is an approach that trains models across multiple clients without centralizing data. 
In FL, each client (e.g., hospitals and institutions) keeps their private data locally and contributes to a shared model by sending updates like gradients or parameters to a central server. This server coordinates the training process, aggregates updates, and broadcasts the refined model back to clients.
The goal of FL is to minimize the global objective function with parameters $\theta$ defined as:
\begin{equation}
    \sum_{k=1}^{K}w_kF_k(\theta) 
    \quad where \quad 
    F_k(w) = \frac{1}{n_k}\sum_{i\in\mathcal{P}_k}l(x_{i}, y_{i}, \theta)
\end{equation}
where $K$ is the number of clients, the weights $w_k$ represents the proportional significance or scale of each local dataset, $n_k$ is the number of training data on client $k$; $\mathcal{P}_k$ is the set of indices of data points on client $k$, and $n_{k}=|\mathcal{P}_k|$; $F_k(\theta)$ is the local objective function; $l(x_{i}, y_{i}, \theta)$ is the loss of the prediction on sample $(x_{i}, y_{i})$.
\par

The traditional centralized FL training process is detailed in Algorithm~\ref{algorithm_fl_basic}, it involves $T$ communication rounds between server and clients.
Specifically, in the $t$-th communication round, each client first downloads the current global model from the server. 
Then each client trains its local model using the local dataset for $E$ local epochs. 
Next, the server collects the model updates of all selected clients and aggregates them into a new global model.
FL training is accomplished by repeating the above round until the global model meets the desired performance criteria.
\par

\begin{figure}[t]
    \centering
    \includegraphics[width=0.7\columnwidth]{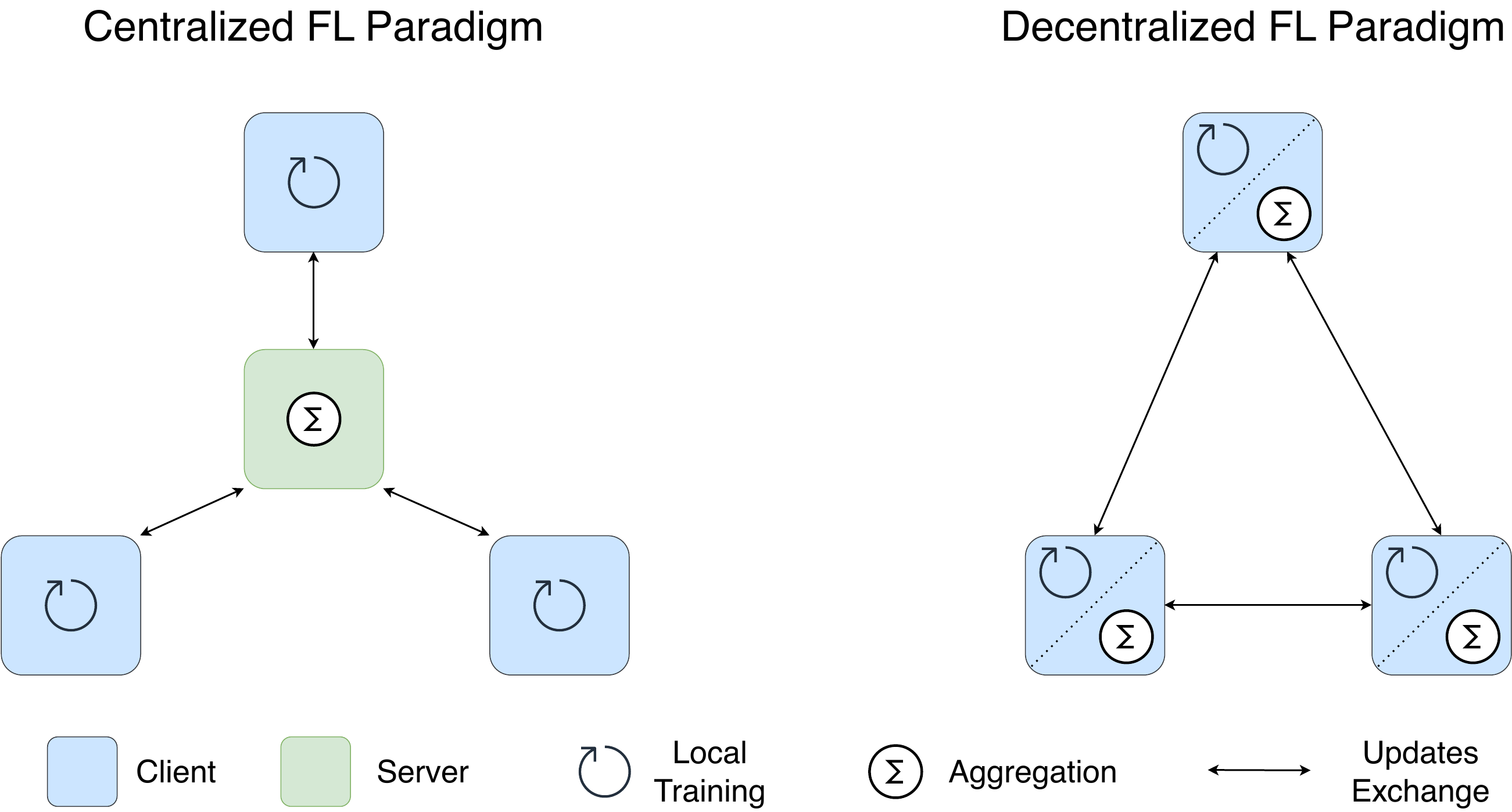}
    \caption{Difference between centralized FL paradigm and decentralized FL paradigm. Centralized FL relies on a central server to manage the training. While decentralized FL eliminates the need for a central server. Instead, clients can directly communicate with connected ones.}
    \label{fig:fl_diff}
\end{figure}

In practice, the rapid development of FL has propelled the field beyond the traditional centralized paradigm, as shown in Figure~\ref{fig:fl_diff}.
For instance, the integration of blockchain~\citep{kumar2021blockchain} and swarm learning~\citep{saldanha2022swarm} has transitioned FL towards decentralized paradigms, such as peer-to-peer, sequential, and cyclic computing, which enhance data privacy, security, and traceability by enabling secure data transactions and consensus mechanisms.
Throughout this evolution, the scope of updates exchanged during communication has expanded.
The updates now encompass not only model parameters or gradients but also partial model parameters~\citep{thapa2022splitfed}, statistical information~\citep{zhu2021data}, and predictions from knowledge distillation techniques, such as logits~\citep{li2023data}.
This expansion helps reduce communication costs, enhance privacy, and enable multi-task learning where only certain parameters are updated collaboratively. 
\par

\begin{figure}[t]
    \centering
    \includegraphics[width=0.7\columnwidth]{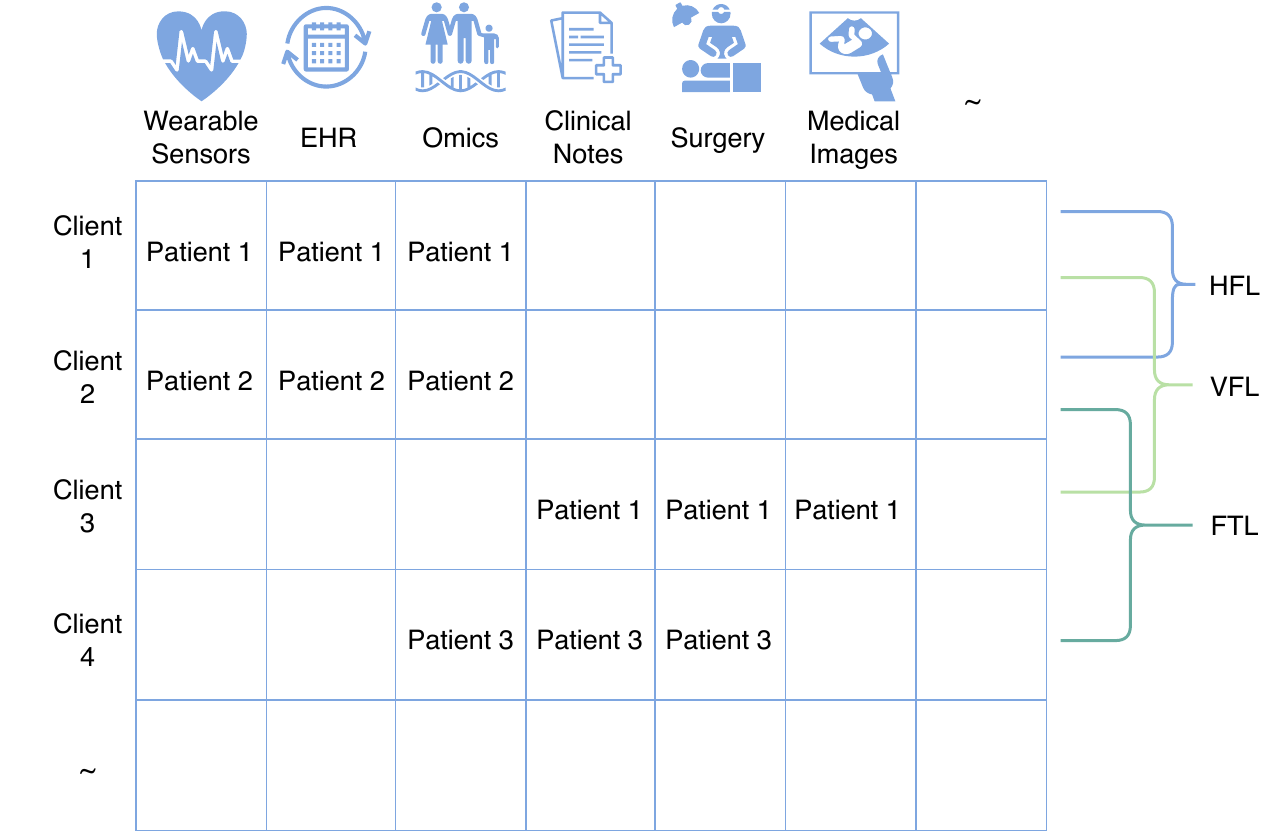}
    \caption{Visual representation of three categories of FL, illustrating their distribution across feature and sample spaces.}
    \label{fig:fl_categories}
\end{figure}

Beyond centralized or decentralized topologies, FL has evolved to address complex scenarios caused by varying feature and sample distributions. This evolution has led to the development of three primary paradigms: \textit{Horizontal Federated Learning (HFL)}, where data from different clients significantly overlap in the feature space but have little overlap in the sample space; \textit{Vertical Federated Learning (VFL)}, where data from different clients have minimal overlap in the feature space but significant overlap in the sample space; and \textit{Federated Transfer Learning (FTL)}, which leverages knowledge transfer to handle scenarios where there is little overlap in both feature and sample spaces. Figure~\ref{fig:fl_categories} illustrates these differences.  
HFL is the most prevalent paradigm in FL studies. For instance, Clients 1 and 2 in Figure~\ref{fig:fl_categories} represent scenarios where a vast number of people use wearables, such as the Apple Watch, to monitor their health conditions. These devices generate large amounts of data that share the same feature space (e.g., heart rate, step count), enabling collaborative model training.  
VFL, on the other hand, is better suited for applications where clients share the same sample space but store distinct features. For example, as illustrated by Clients 1 and 3 in Figure~\ref{fig:fl_categories}, a patient’s medical records may be distributed across multiple hospitals, with each hospital contributing unique features (e.g., imaging data, lab results). Aggregating these features allows for a more comprehensive model.  
Finally, FTL addresses scenarios with limited overlap in both feature and sample spaces. As depicted by Clients 2 and 4 in Figure~\ref{fig:fl_categories}, different clients may manage varying combinations of healthcare data and patient populations, with only a small intersection in the feature space. This approach is particularly relevant for tabular EHR data~\citep{liu2022confederated}.
\par


FL can be further categorized into \textit{Cross-silo FL} and \textit{Cross-device FL} based on the scale and attributes of participating clients.
\textit{Cross-silo FL} is tailored for scenarios where a limited number of participating clients, such as hospitals, medical centres, and institutions, collaboratively engage in all stages of FL training. Notable examples include Healthchain~\citep{healthchain}, which facilitates FL deployment among multiple hospitals in Europe, and the Melody Project~\citep{melody}, designed to optimize the drug discovery task across multiple companies while preserving data privacy.
\textit{Cross-device FL}, on the other hand, is designed for scenarios involving a multitude of participating clients, typically edge devices with limited data storage and computing capabilities. Examples include wearables (e.g., Apple Watch) and Internet of Medical Things (IoMT) devices. For instance, IoMT devices like Raspberry Pi and Jetson Nano can collect electronic health records (EHRs) in resource-limited environments, enabling early detection of sepsis~\citep{alam2023fedsepsis}.
\par

\section{Method}

Our review was conducted in accordance with the Preferred Reporting Items for Systematic Reviews and Meta-Analyses (PRISMA) guidelines~\citep{moher2009preferred}.
As shown in Figure~\ref{fig:fl_prisma}, the flow diagram outlines the search, inclusion and exclusion procedures.
We carried out a comprehensive search of the most recent studies focusing on advanced FL technologies within healthcare domain up to May 2024.

\subsection{Literature Search}
We conducted a systematic search using PubMed, Web of Science, Scopus, Science Direct, IEEEXplore, ArXiv, Springerlink, ACM Digital Library, and Google Scholar.
Any study up to May 2024 that involved the use of FL technologies in healthcare based on a simulated or real distributed scenario was included.
The search phrases included the following keywords: ``Federate Learning", ``Healthcare", ``Privacy-Preserving", ``Medical", ``Biomedical", ``Decentralized Learning", and ``Privacy", using Boolean operators such as ``AND/OR" and various combinations of these keywords.
As shown in Figure~\ref{fig:fl_prisma}, initially, a total of 1,149 studies were identified.

\subsection{Study Selection}
We defined clear and transparent inclusion and exclusion criteria as follows.
\textit{Inclusion criteria:} 
(1) Studies involving the implementation of FL in the healthcare domain;
(2) Studies that, while not explicitly focused on healthcare applications, involve or utilize healthcare data or scenarios in their experiments;
(3) Studies on the design or optimization of FL frameworks/workflows that cover the healthcare domain;
(4) Studies in English language.
\textit{Exclusion criteria:} 
(1) Duplicate studies;
(2) Studies such as surveys, reviews, opinions, editorial letters, book chapters, and theses;
(3) Studies unrelated to FL or those using FL for non-healthcare applications;
(4) Non-English language studies.

\begin{figure}[t]
    \centering
    \includegraphics[width=0.6\columnwidth]{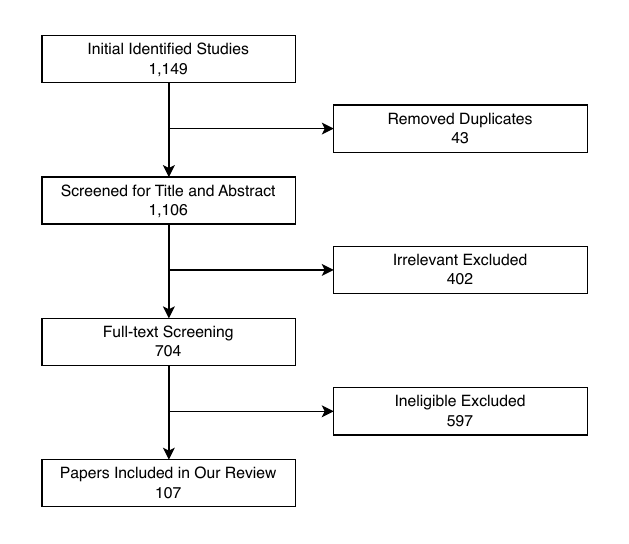}
    \caption{PRISMA flow diagram for our review, highlighting the inclusion and exclusion of studies at each stage.}
    \label{fig:fl_prisma}
\end{figure}

Based on the above criteria, the screening procedure was conducted independently by two groups of authors (Group A: M.L. and P.X.; Group B: J.H. and Z.T.) to eliminate bias and ambiguity. 
Two groups confirmed the selected studies and resolved any conflicts or inconsistencies through discussion between the groups.
The study selection process is outlined in Figure~\ref{fig:fl_prisma}. 
Initially, a total of 43 duplicate studies were removed. 
Subsequently, the titles and abstracts were carefully screened, leading to the exclusion of 402 unqualified and irrelevant studies.
Next, the eligibility of the remaining 704 studies was assessed through full-text screening.
Finally, after further evaluation, 597 studies were deemed ineligible, and 107 studies were included in our final review.

\section{Results}

%
\subsection{Application and Data}
Included studies explored a broad range of healthcare specialties, including general medicine~\citep{sheller2020federated}, cardiology~\citep{linardos2022federated}, oncology~\citep{chakravarty2021federated}, ophthalmology~\citep{liu2021feddg}, drug discovery~\citep{melody}, multiomics~\citep{zhou2024ppml}, dermatology~\citep{haggenmuller2024federated}, and radiology~\citep{malik2023dmfl_net}. 
Most studies focused on tasks such as classification (67/107), segmentation (20/107), and detection (8/107), with additional applications in regression~\citep{sadilek2021privacy}, clustering~\citep{zhou2024ppml}, reconstruction~\citep{yan2024cross,zou2023self}, feature selection~\citep{sun2021fedio,lu2022federated}, data synthesis~\citep{wang2023fedmed,dalmaz2024one,jin2023backdoor}, and biomedical language process~\citep{peng2024depth}.
In terms of data types, medical imaging, including MRI, CT, and X-rays, was the most frequently used (55/107), followed by clinical data and electronic health records (EHR) (24/107), skin images (6/107), retinal images (6/107), histopathology slides (16/107), multiomics data (3/107), and biomedical language data (1/107). Some studies involved multiple data types, while 8 studies did not specify the type of data used or used non-healthcare data~\citep{tian2023robust,li2022contract,wang2023privacy}.

\begin{table*}[h!]
\begin{center}
    \caption{Key results of included studies.}
    \label{table:table_summary}
    \resizebox{!}{115mm}{
    \begin{tabular}{l m{100mm} m{40mm} m{70mm}}
    \toprule
    Item & \makecell[l]{Characteristics} & \makecell[l]{Number of Study} & \makecell[l]{Examples} \\
    \midrule    
    Cohort Size & 
        \begin{item} 
            \item $\leq100$ 
            \item $100-1000$ 
            \item $\geq1000$ 
            \item unavailable
        \end{item}  &
        \begin{item} 
            \item 6
            \item 9
            \item 24
            \item 68
        \end{item} & 
        \begin{item} 
            \item \cite{kumar2021blockchain,kalapaaking2022smpc,che2022federated}
            \item \cite{sheller2019multi,sheller2020federated,ogier2023federated} 
            \item \cite{dayan2021federated,sadilek2021privacy,fets} 
            \item \cite{zhang2023grace,madni2023federated,yan2024cross} 
        \end{item} \\
    \hline
    
    Task & 
        \begin{item} 
            \item Classification 
            \item Segmentation 
            \item Detection 
            \item Others
        \end{item} &
        \begin{item} 
            \item 74
            \item 22
            \item 8
            \item 10
        \end{item} &
        \begin{item} 
            \item \cite{gong2021ensemble,ogier2022flamby,zhang2024unified}
            \item \cite{souza2021multi,tedeschini2022decentralized,lin2023unifying} 
            \item \cite{nguyen2021federated,alam2023fedsepsis,baheti2020federated} 
            \item \cite{sadilek2021privacy,zhou2024ppml,zou2023self} 
        \end{item} \\
    \hline

    Data Type &
        \begin{item} 
            \item Medical Imaging (MRI, CT, X-rays) 
            \item Clinical and EHR 
            \item Skin Images
            \item Retinal Images
            \item Histopathology Slides
            \item Multiomics
            \item Biomedical Language
            \item unavailable or non-healthcare
        \end{item} &
        \begin{item} 
            \item 55
            \item 24
            \item 6
            \item 6
            \item 16
            \item 3
            \item 1
            \item 8
        \end{item} &
        \begin{item} 
            \item \cite{zhang2021dynamic,sheller2020federated,dayan2021federated} 
            \item \cite{sakib2021asynchronous,kerkouche2021privacy,soltan2024scalable} 
            \item \cite{zhang2023grace,lian2022deep,zhang2022homomorphic} 
            \item \cite{hatamizadeh2023gradient,lian2022deep,lin2023unifying} 
            \item \cite{saldanha2022swarm,truhn2024encrypted,jiang2022harmofl}
            \item \cite{zhou2024ppml,warnat2021swarm,froelicher2021truly}
            \item \cite{peng2024depth}
            \item \cite{repetto2022federated,wang2023privacy,li2022contract} 
        \end{item} \\
    \hline

    Number of Clients & 
        \begin{item} 
            \item $\leq10$ 
            \item $10-50$ 
            \item $\geq50$ 
            \item unavailable 
        \end{item} &
        \begin{item} 
            \item 57
            \item 17
            \item 6
            \item 27
        \end{item} &
        \begin{item} 
            \item \cite{ogier2023federated,nvidia2020,melody}
            \item \cite{sheller2019multi,dayan2021federated,tfda2024}
            \item \cite{kerkouche2021privacy,repetto2022federated,balkus2022federated}
            \item \cite{kalapaaking2023blockchain,salim2024articlesfederated,wang2023privacy}
        \end{item} \\
    \hline

    Topology &
        \begin{item} 
            \item Centralized
            \item Decentralized
        \end{item} &
        \begin{item} 
            \item 95
            \item 12
        \end{item} &
        \begin{item} 
            \item \cite{gong2021ensemble,ogier2022flamby,yan2024cross} 
            \item \cite{rehman2022secure,chang2021blockchain,tian2023robust}
        \end{item} \\
    \hline

    Type of Federation & 
        \begin{item} 
            \item Cross-Silo 
            \item Cross-Device 
            \item unavailable 
        \end{item} &
        \begin{item} 
            \item 98
            \item 7
            \item 2
        \end{item} &
        \begin{item} 
            \item \cite{bercea2021feddis,zou2023self,lin2023unifying} 
            \item \cite{chang2021blockchain,rehman2022secure,chen2020fedhealth}
            \item \cite{wang2023privacy,salim2024articlesfederated}
        \end{item} \\
    \hline

    Scenario & 
        \begin{item} 
            \item Deployment 
            \item Simulation
        \end{item} &
        \begin{item} 
            \item 10
            \item 97
        \end{item} &
        \begin{item} 
            \item \cite{mullie2024coda,roth2022nvidia,cremonesi2023fed} 
            \item \cite{chen2020achieving,li2021federated,zhang2022homomorphic} 
        \end{item} \\
    \hline

    Framework & 
        \begin{item} 
            \item Custom-designed
            \item Open-source Options
            \item unavailable
        \end{item} &
        \begin{item} 
            \item 78
            \item 13
            \item 16
        \end{item} &
        \begin{item} 
            \item \cite{li2021federated,chen2020achieving,sav2022privacy}
            \item \cite{dragon,dayan2021federated,roth2022nvidia} 
            \item \cite{madni2023federated,zhang2022homomorphic,bey2020fold} 
        \end{item} \\
    \hline

    Data Curation &
        \begin{item}
            \item Standardization \& Harmonization
            \item unavailable
        \end{item} &
        \begin{item}
            \item 12
            \item 95
        \end{item} &
        \begin{item}
            \item \cite{ogier2023federated,truhn2024encrypted,kumar2021blockchain}
            \item \cite{zhang2023grace,rehman2024fedcscd,lakhan2024digital}
        \end{item} \\
    \hline

    \multirow{4}{*}{Data Partition} &
        \begin{item}
            \item HFL
            \item VFL
            \item FTL
            \item unavailable
        \end{item} &
        \begin{item}
            \item 94
            \item 3
            \item 1
            \item 12
        \end{item} &
        \begin{item}
            \item \cite{zhou2024distributed,yaqoob2023federated,feng2024robustly}
            \item \cite{liu2022confederated,che2022federated,yan2024cross}
            \item \cite{chen2020fedhealth}
            \item \cite{wang2023fedmed,andreux2020federated,lu2022federated}
        \end{item} \\
    \cline{2-4} &
        \begin{item}
            \item Natural Split
            \item Simulate
            \item unavailable
        \end{item} &
        \begin{item}
            \item 37
            \item 67
            \item 3
        \end{item} &
        \begin{item}
            \item \cite{chang2021blockchain,tong2022distributed,zou2023self}
            \item \cite{zhou2024federated,peng2024depth,lakhan2024digital}
            \item \cite{mateus2024data,roth2022nvidia,tfda2024}
        \end{item} \\
    \cline{2-4} &
        \begin{item}
            \item Train/Test/Val Details per Client
        \end{item} &
        \begin{item}
            \item 17
        \end{item} &
        \begin{item}
            \item \cite{peng2024depth,feng2024robustly,mazher2024self}
        \end{item} \\
    \cline{2-4} &
        \begin{item}
            \item Holdout Cohort for Evaluation
        \end{item} &
        \begin{item}
            \item 12
        \end{item} &
        \begin{item}
            \item \cite{zhang2023grace,zou2023self,truhn2024encrypted}
        \end{item} \\
    \hline

    \multirow{2}{*}{Model} &
        \begin{item}
            \item Deep Learning
            \item Traditional Machine Learning
            \item unavailable
        \end{item} &
        \begin{item}
            \item 83
            \item 7
            \item 17
        \end{item} &
        \begin{item}
            \item \cite{mullie2024coda,cremonesi2023fed,alam2023fedsepsis}
            \item \cite{ogier2023federated,sadilek2021privacy,soltan2024scalable}
            \item \cite{ma2021communication,dragon,roth2022nvidia}
        \end{item} \\
    \cline{2-4} &
        \begin{item}
            \item Initialization: Random
            \item Initialization: Pretrained/Foundation Models
            \item unavailable
        \end{item} &
        \begin{item}
            \item 18
            \item 5
            \item 84
        \end{item} &
        \begin{item}
            \item \cite{ogier2023federated,liu2022confederated,hatamizadeh2023gradient}
            \item \cite{li2024begin,haggenmuller2024federated,peng2024depth}
            \item \cite{wang2022privacy,sav2022privacy,zhang2024unified}
        \end{item} \\
    \hline

    \multirow{6}{*}{Optimization} &
        \begin{item}
            \item System Heterogeneity
        \end{item} &
        \begin{item}
            \item 0
        \end{item} &
        \begin{item}
            \item -
        \end{item} \\
    \cline{2-4} &
        \begin{item}
            \item Generalization in Open Domain
        \end{item} &
        \begin{item}
            \item 15
        \end{item} &
        \begin{item}
            \item \cite{ogier2023federated,liu2022confederated,haggenmuller2024federated} 
        \end{item} \\
    \cline{2-4} &
        \begin{item}
            \item Communication Efficiency
        \end{item} &
        \begin{item}
            \item 19
        \end{item} &
        \begin{item}
            \item \cite{kerkouche2021privacy,li2023data,qu2022handling}
        \end{item} \\
    \cline{2-4} &
        \begin{item}
            \item Theoretical Convergence Analysis
        \end{item} &
        \begin{item}
            \item 0
        \end{item} &
        \begin{item}
            \item -
        \end{item} \\
    \cline{2-4} &
        \begin{item}
            \item Temporal Data Dynamics 
        \end{item} &
        \begin{item}
            \item 2
        \end{item} &
        \begin{item}
            \item \cite{alam2023fedsepsis,liu2022confederated}
        \end{item} \\
    \cline{2-4} &
        \begin{item}
            \item Synchronous Aggregation
            \item Asynchronous Aggregation
        \end{item} &
        \begin{item}
            \item 92
            \item 15
        \end{item} &
        \begin{item}
            \item \cite{nguyen2021federated,kumar2021blockchain,saldanha2022swarm}
            \item \cite{zhou2024ppml,haggenmuller2024federated,malik2023dmfl_net}
        \end{item} \\        
    \hline

    Privacy and Security &
        \begin{item}
            \item Model Updates Protection
        \end{item} &
        \begin{item}
            \item 41
        \end{item} &
        \begin{item}
            \item \cite{nguyen2021federated,xie2024mhpflid,zhang2022homomorphic} 
        \end{item} \\
    \hline

    Open Source &
        \begin{item} 
            \item Code Available 
            \item Trained Model Available 
            \item unavailable 
        \end{item} &
        \begin{item} 
            \item 29
            \item 1
            \item 78
        \end{item} &
        \begin{item} 
            \item \cite{gong2021ensemble,ogier2022flamby,hatamizadeh2023gradient} 
            \item \cite{dayan2021federated}
            \item \cite{li2022integrated,kalapaaking2022smpc,qu2022handling}
        \end{item} \\
    \bottomrule
    \end{tabular}}
\end{center}
\end{table*}

%
\subsection{Topology, Scenario and Framework}
The centralized FL paradigm dominates current implementations, with 95 out of 107 studies following this topology. 
Only 10 studies reported real-world deployments in distributed clinical settings, while the rest remained in the realm of prototypes or simulations. 
In terms of frameworks, the majority (78/107) utilized custom-designed FL frameworks, while a smaller number (13/107) employed open-source options such as Flower~\citep{dragon}, Flare~\citep{dayan2021federated,roth2022nvidia,nvidia2020}, SubstraFL~\citep{ogier2023federated,melody}, TFF~\citep{sadilek2021privacy}, OpenFL~\citep{fets}, PySyft~\citep{malik2023dmfl_net,kalapaaking2022smpc,gawali2021comparison,truhn2024encrypted}, and FedBioMed~\citep{cremonesi2023fed}. 
16 studies did not specify the framework used. 
Further details about open-source frameworks can be found in Table~\ref{table:fl_tools}.

%
\subsection{Data Curation and Partition}
Among the reviewed studies, only 12 provided details on the processes of data standardization and harmonization.
Regarding data partition, HFL was the predominant approach, with 94 out of 107 studies focusing solely on it. 
In contrast, VFL was explored in only 3 studies~\citep{yan2024cross,liu2022confederated,che2022federated}, while 2 studies considered both HFL and VFL in combination~\citep{liu2022confederated,che2022federated}. 
Only 1 study discussed FTL~\citep{chen2020fedhealth}.
Notably, 12 studies did not mention this aspect at all.
The majority of studies addressed only one type of data heterogeneity, such as quantity skew or label skew, without considering multiple factors simultaneously. Moreover, 37 studies employed natural data splits for training and/or evaluation, while the rest relied on artificial splits.
Only 17 studies detailed their training, testing or validation sets and 12 studies split a holdout cohort for evaluation.

%
\subsection{Model}
Among reviewed studies, Convolutional Neural Networks (CNNs) were the most commonly utilized (80/107), including both custom models specifically designed for healthcare tasks and well-established architectures like ResNet, DenseNet, MobileNet, and U-Net~\citep{li2022integrated,kalapaaking2022smpc,qu2022handling}. 
Additionally, Recurrent Neural Networks (RNNs) have been incorporated to leverage their strengths in handling complex healthcare data~\citep{che2022federated,alam2023fedsepsis,paragliola2022definition}. 
Some studies also employed custom Multi-Layer Perceptrons (MLPs) and attention mechanisms to further boost model performance~\citep{repetto2022federated,kandati2022genetic,kerkouche2021privacy}.
\cite{peng2024depth} utilized large language models for distributed biomedical natural language processing.
Beyond deep learning approaches, several studies explored traditional machine learning (ML) algorithms, including linear models and ensemble methods. Notable examples include logistic regression~\citep{repetto2022federated}, support vector machines~\citep{bey2020fold}, fuzzy clustering~\citep{brisimi2018federated}, and decision trees~\citep{balkus2022federated,aminifar2022extremely}.
Interestingly, only 23 studies explicitly discussed their initialization strategies for model training.
Among these, the majority opted for random initialization, while a mere five clearly stated that they utilized pretrained or foundation models as their starting point~\citep{li2024begin,adnan2022federated,haggenmuller2024federated,linardos2022federated,peng2024depth}.

%
\subsection{Optimization}
Most studies addressed either data or model heterogeneity, and none of them considered system heterogeneity. Only 15 studies evaluated model generalization ability in unseen open domains. 
A total of 19 studies focused on improving communication efficiency, employing techniques such as knowledge distillation~\citep{gong2021ensemble,xie2024mhpflid,li2024begin}, gradient quantization~\citep{kerkouche2021privacy}, one-shot FL~\citep{qu2022handling}, split learning~\citep{gawali2021comparison}, and tensor factorization~\citep{ma2021communication,maj2021communication}.
In terms of convergence analysis, a few studies (21/107) reported metrics such as communication rounds and costs, as well as overall convergence time, but none provided a theoretical understanding of convergence dynamics. 
Only two studies considered temporal data dynamics in model learning~\citep{liu2022confederated,alam2023fedsepsis}.
Regarding synchronization, 15 studies employed asynchronous aggregation instead of synchronous aggregation, particularly in applications involving wearables~\citep{chang2021blockchain} and IoMT devices~\citep{alam2023fedsepsis}.

%
\subsection{Privacy and Security}
Only 41 studies addressed the exchange of model updates with privacy guarantees. The most commonly used techniques for safeguarding model updates included Differential Privacy (DP), Homomorphic Encryption (HE), Secure Multi-Party Computation (SMPC), knowledge distillation, and partial model exchange. However, metadata such as sample sizes and distributions were frequently shared without protection, particularly in methods based on FedAvg~\citep{adnan2022federated,ziller2021differentially}.
To mitigate the risk of adversaries inferring raw data, synthetic data was employed in some cases~\citep{nguyen2021federated,qu2022handling,jin2023backdoor,rehman2024fedcscd}. Additionally, swarm learning and blockchain were utilized to secure the communication process~\citep{saldanha2022swarm,nguyen2021federated}.

%
\subsection{Fairness and Incentive}
Only three studies have discussed issues related to fairness and/or incentives in healthcare FL~\citep{hosseini2023proportionally,li2022contract,zhang2024unified}, with just one of these studies specifically exploring the complexities of both fairness and incentives in detail~\citep{zhang2024unified}.
\par
In the context of FL for healthcare, fairness generally refers to the equitable distribution of model performance among participants, ensuring that no entity is disadvantaged. Incentives are mechanisms designed to motivate healthcare institutions to participate in federated networks, often by offering rewards for contributions such as high-quality data or computational resources. For a more comprehensive discussion of fairness and incentive in healthcare FL, we refer readers to Section~\ref{sec5-fair-incentive}, where these concepts are explored in greater detail.

%
\subsection{Evaluation}
Most studies used conventional ML metrics for evaluation, such as accuracy, precision, area under the receiver operating characteristic curve (AUC), sensitivity/recall, specificity, F1-score, Dice score, Intersection over Union (IoU), Hausdorff Distance (HD), and loss value. Additionally, many studies performed comparisons against classical centralized models or localized models, and conducted ablation studies.
However, only a few studies (26/107) addressed critical aspects unique to FL, such as communication overhead, resource consumption, scalability, generalization, privacy, fairness, and security concerns. 
As for benchmarking, just one study provided relatively comprehensive benchmarks across multiple healthcare datasets~\citep{ogier2022flamby}.
Interpretability was explored in seven studies, either through feature selection~\citep{soltan2024scalable,sun2021fedio}, attention maps~\citep{lu2022federated,gong2021ensemble,truhn2024encrypted}, or tree-based models~\citep{ogier2023federated,sadilek2021privacy}. While 29 studies released their source code, only one also made the trained model publicly available~\citep{dayan2021federated}.

We provide a more detailed summary of the key results in Table~\ref{table:table_summary}.
%

\begin{figure*}[ht!]
    \centering
    \includegraphics[width=\linewidth]{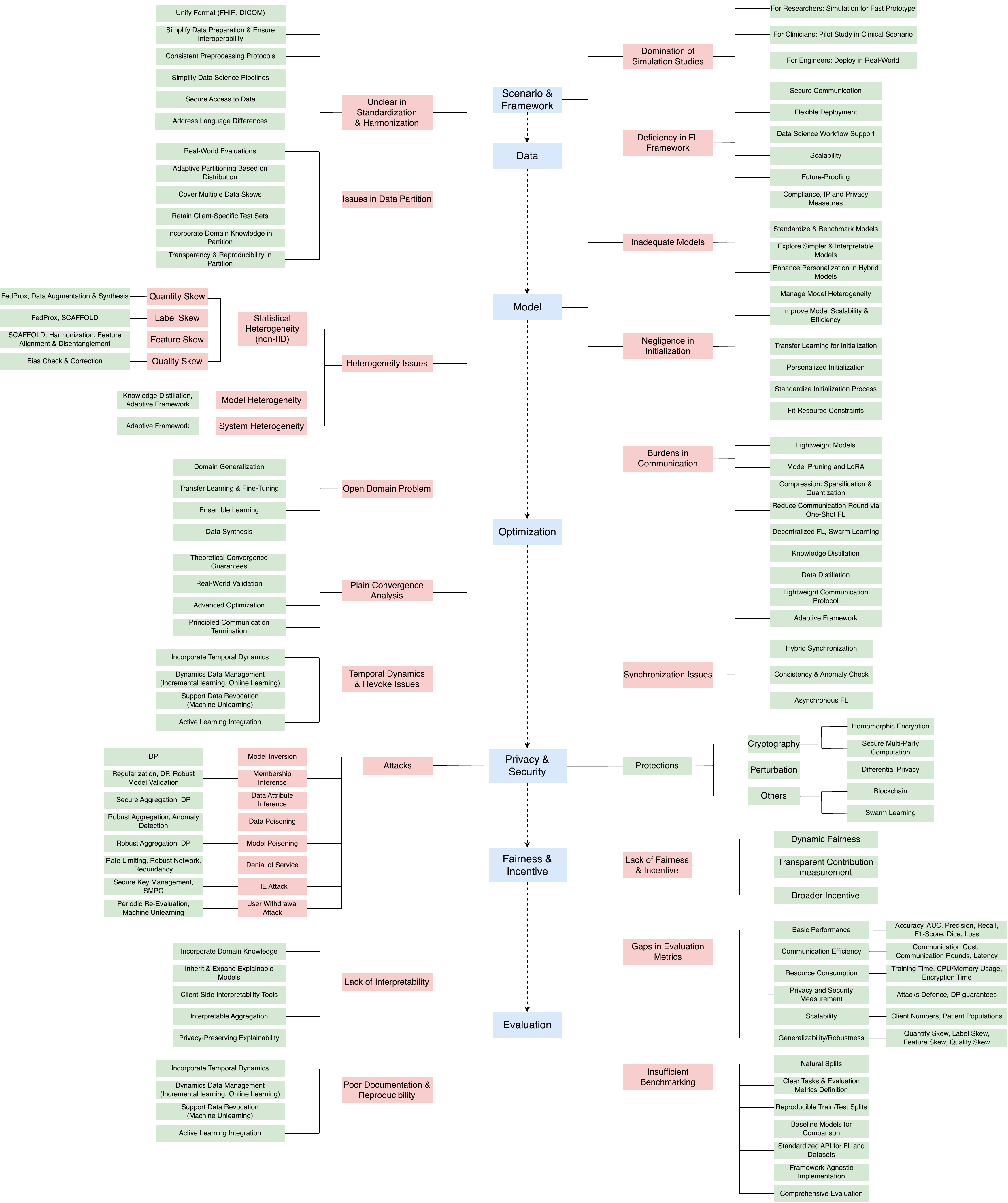}
    \caption{Taxonomy of challenges and pitfalls (red blocks) as well as recommended solutions and opportunities (green blocks).}
    \label{fig:fl_tree}
\end{figure*}

\section{From Challenges and Pitfalls to Recommended Solutions and Future Opportunities}
After a thorough review of the most recent and advanced FL studies, we find various challenges and pitfalls that still limit the implementation of FL in healthcare.
In this section, we introduce a clear taxonomy, as depicted in Figure~\ref{fig:fl_tree}, focusing on the challenges and pitfalls, and further providing recommended solutions and opportunities.
We adhere to the best practice workflow in FL for discussion in the following subsections.

\begin{figure}[t]
    \centering
    \includegraphics[width=0.6\columnwidth]{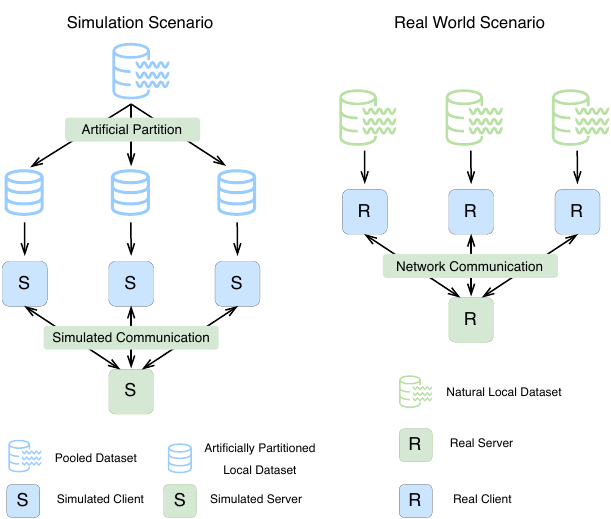}
    \caption{Simulation scenario VS. real-world distributed scenario.}
    \label{fig:fl_simulationVSdeploy}
\end{figure}

\subsection{Scenario and Framework}
\subsubsection{Domination of Simulation Studies}
The majority of existing studies have been confined to simulation environments, with only 10 studies incorporating real-world distributed clinical scenarios. This indicates that the application of FL in healthcare is still in its nascent stage. 
The complexity of deploying FL across a real-world network of hospitals and institutions has significantly hindered its progress. 
Most studies have operated within controlled, simulated settings where data is pooled and then artificially partitioned to represent distributed environments. The simulated clients interact with a simulated server, coordinating model updates in a manner that is highly controlled and predictable.
In contrast, real-world scenarios involve each client working with inherently distributed, heterogeneous, and locale-specific data. The interaction between real server and clients is far more complex, requiring secure protocols, real-time communication, and the ability to handle diverse datasets across various institutions. This disparity between simulation and real-world environments is illustrated in Figure~\ref{fig:fl_simulationVSdeploy}.
\par
Moreover, very few studies have explored FL practices at a national or international scale~\citep{dayan2021federated,soltan2024scalable}.
Notable examples include the Collaborative Data Analysis (CODA)~\citep{mullie2024coda}, \cite{melody}, HealthChain~\citep{healthchain} and DRAGON~\citep{dragon}. 
CODA tested FL’s feasibility across eight hospitals in Canada by enrolling patients with suspected or confirmed COVID-19 over three years.
Melody deployed multi-task FL among 10 pharmaceutical companies to optimize the drug discovery process.
HealthChain and DRAGON implemented FL across multiple hospitals in Europe, facilitating the prediction of treatment responses for cancer and COVID-19 patients.
\par
Despite these rare promising examples, the majority of FL studies remain proof-of-concept and the broader deployment of FL in healthcare remains largely undocumented. There is still a lack of clarity on how FL nodes are set up within individual hospitals, the methods for delivering local models to these nodes, the protocols enabling interaction between nodes and aggregators, and the mechanisms triggering new training rounds, etc.

\paragraph{\textbf{Recommendations \& Opportunities}}
\begin{itemize}
    \item \textit{For researchers}, it is recommended to continue leveraging simulation environments to rapidly prototype and evaluate FL algorithms. Simulations offer precise control over experimental conditions, which is essential for understanding the underlying mechanics of FL and its behaviour under various scenarios. However, researchers should acknowledge the limitations of simulations and anticipate the challenges that real-world deployments may introduce. Beyond simulations, to enhance data diversity, collaboration and generalization, efforts should be made to implement FL at a national and international scale, with cloud computing offering scalable resources and seamless implementation across institutions.
    \item \textit{For clinicians} interested in applying FL to enhance clinical diagnostics and prognostics, it is crucial to comprehend both the potential benefits and limitations of FL. Clinicians should collaborate closely with researchers and engineers to identify promising use cases for FL in clinical practice. This may involve conducting pilot studies to assess the feasibility and effectiveness of FL in specific clinical scenarios. Clinicians should advocate for the integration of FL into existing healthcare workflows to ensure a seamless transition from research to practice. Additionally, their feedback on the usability and impact of FL systems is vital for guiding further refinements.
    \item \textit{For engineers}, the focus should be on addressing the practical challenges of FL deployment in real-world settings. This includes ensuring the interoperability of different hospital systems, safeguarding data privacy and security, and managing the communication overhead of networks. Engineers should aim to develop robust and scalable solutions adaptable to the heterogeneous IT infrastructures across healthcare institutions. Close collaboration with clinicians and researchers is essential to ensure that FL systems meet healthcare-specific needs and comply with regulatory standards.
\end{itemize}

\begin{table*}[t]
    \centering
    \caption{Capabilities and features of current popular FL frameworks.}
    \resizebox{!}{25mm}{
    \begin{tabular}{lccccccccc}
    \toprule
    Framework & Developer & 
    \makecell[c]{Secure\\Aggregation} &  
    \makecell[c]{Communication\\Efficiency} & 
    \makecell[c]{Healthcare\\Adaptation} & 
    Traceability & 
    Deployment & 
    \makecell[c]{Foundation\\Model} & 
    Scalability & 
    \makecell[c]{Clound\\Friendly}\\
    \midrule
    Flare~\citep{roth2022nvidia} & Nvidia 
    & DP, HE & - & - & \checkmark & \checkmark & \checkmark & \checkmark & \checkmark \\
    FedML~\citep{he2020fedml} & TensorOpera 
    & DP, HE & \checkmark & - & \checkmark & \checkmark & \checkmark & \checkmark & \checkmark \\
    FederatedScope~\citep{xie2022federatedscope} & Alibaba 
    & DP, SMPC & - & - & - & - & \checkmark & - & - \\
    Flower~\citep{beutel2022flower} & FlowerLab 
    & DP & - & - & \checkmark & \checkmark & \checkmark & \checkmark & \checkmark \\
    FATE~\citep{liu2021fate} & WeBank 
    & DP, HE, SMPC & - & - & \checkmark & \checkmark & \checkmark & \checkmark & \checkmark \\
    SubstraFL~\citep{substrafl2024} & Owkin 
    & DP, SMPC & - & \checkmark & \checkmark & \checkmark & - & \checkmark & \checkmark \\
    PySyft~\citep{ryffel2018generic} & OpenMined 
    & DP & - & - & - & - & - & - & - \\
    OpenFL~\citep{foley2022openfl} & Intel 
    & DP & - & - & - & \checkmark & - & - & - \\
    TFF~\citep{bonawitz2019towards} & Google 
    & DP & - & - & - & \checkmark & - & - & - \\
    Fed-BioMed~\citep{cremonesi2023fed} & Inria 
    & DP & - & \checkmark & - & \checkmark & - & - & - \\
    IBM FL~\citep{ludwig2020ibm} & IBM 
    & DP, SMPC & - & - & - & \checkmark & - & - & - \\
    PaddleFL~\citep{ma2019paddlepaddle} & Baidu 
    & DP, SMPC & - & - & - & \checkmark & - & - & - \\
    SAFEFL~\citep{gehlhar2023safefl} & ENCRYPTO
    & SMPC & - & - & - & - & - & - & - \\
    \bottomrule
    \end{tabular}}
    \label{table:fl_tools}
\end{table*}

\subsubsection{Deficiency in FL Frameworks Development and Usage}
Most studies developed their own FL frameworks, often not strictly aligning with standard FL protocols, particularly when confined to single machine simulation studies. 
Meanwhile, some simulation studies used industrial-grade frameworks, which introduce unnecessary complexity and resource demands for simulation and prototype research.
Some other studies utilized lightweight open-source FL frameworks, although prevalent, frequently lack healthcare-specific adaptations, leading to deficiencies in privacy, security, and regulatory compliance. 
Common shortcomings across current FL frameworks include a lack of healthcare adaptations, as most frameworks are not tailored to meet healthcare-specific requirements, which include stringent privacy, security, and regulatory standards. 
Additionally, many frameworks do not address the need for communication efficiency, which is essential for the practical deployment of FL in resource-constrained environments. 
Limited support for traceability also hinders accountability and transparency in FL. 
Furthermore, while some frameworks offer scalability and cloud compatibility, many do not, which can limit their ability to handle large-scale healthcare data and integrate with existing cloud infrastructures. 
Here, we inventory the most popular FL frameworks in Table~\ref{table:fl_tools}, with emphasis on those adapted for healthcare, and outline their features.
\par

\paragraph{\textbf{Recommendations \& Opportunities}}
\begin{itemize}
    \item \textit{For researchers} aiming to swiftly prototype and test novel concepts can benefit from frameworks that incorporate comprehensive simulator modules. These tools allow for the rapid iteration and validation of ideas within a controlled simulation environment, which can be critical for the initial stages of research and development.
    \item \textit{For engineers} seeking to deploy FL in real-world scenarios should consider frameworks tailored to the specific needs of the healthcare domain. These frameworks should offer healthcare specific adaptations to ensure compatibility with medical data formats, regulatory compliance, and the unique challenges of healthcare data analysis.
    \item \textit{Users} facing computational constraints are encouraged to explore cloud-friendly frameworks that leverage cloud computing services such as Azure and AWS. These platforms can alleviate the burden of substantial computational demands and the complexities of local infrastructure development. Moreover, cloud computing can significantly mitigate the risk of network issues that may arise from client-hosted infrastructures with varying capabilities.
    \item \textit{More specifically,}  we propose the following suggestions for FL framework selection, usage, and development:
    \begin{itemize}
        \item \textit{Secure Communication:} The integrity of the FL system hinges on secure communication protocols, where encryption should be employed~\citep{li2023review}.
        \item \textit{Flexible Deployment:} To streamline the deployment process, FL frameworks should support secure, reliable, and flexible deployment methods. They should integrate seamlessly with existing IT and data science infrastructures, facilitating a routine and uneventful deployment experience~\citep{lo2022architectural}.
        \item \textit{Data Science Workflow Support:} Given the necessity for diverse data providers to achieve robust FL outcomes, frameworks should support a comprehensive data science workflow. The ideal framework should be agnostic to both the model and the data, accommodating a wide range of data types and analytical methods~\citep{cremonesi2023fed}.
        \item \textit{Scalability:} Scalability is a key consideration for FL frameworks, which must accommodate an increasing number of participants and the corresponding complexity. Addressing scalability challenges, particularly with privacy-enhancing technologies such as synthetic data or HE, is crucial for the long-term viability of FL initiatives~\citep{lai2022fedscale}.
        \item \textit{Future-proofing:} FL frameworks should be designed with future-proofing in mind, anticipating emerging use cases, evolving security threats, and new privacy concerns. It should facilitate the dynamic participation of data providers, adapt computational resources to fluctuating client numbers, and implement regular system updates to address privacy and security challenges~\citep{lo2022architectural}.
    \end{itemize}
\end{itemize}

\subsection{Data}
\subsubsection{Unclear in Data Standardization and Harmonization}
\begin{figure}[t]
    \centering
    \includegraphics[width=0.6\columnwidth]{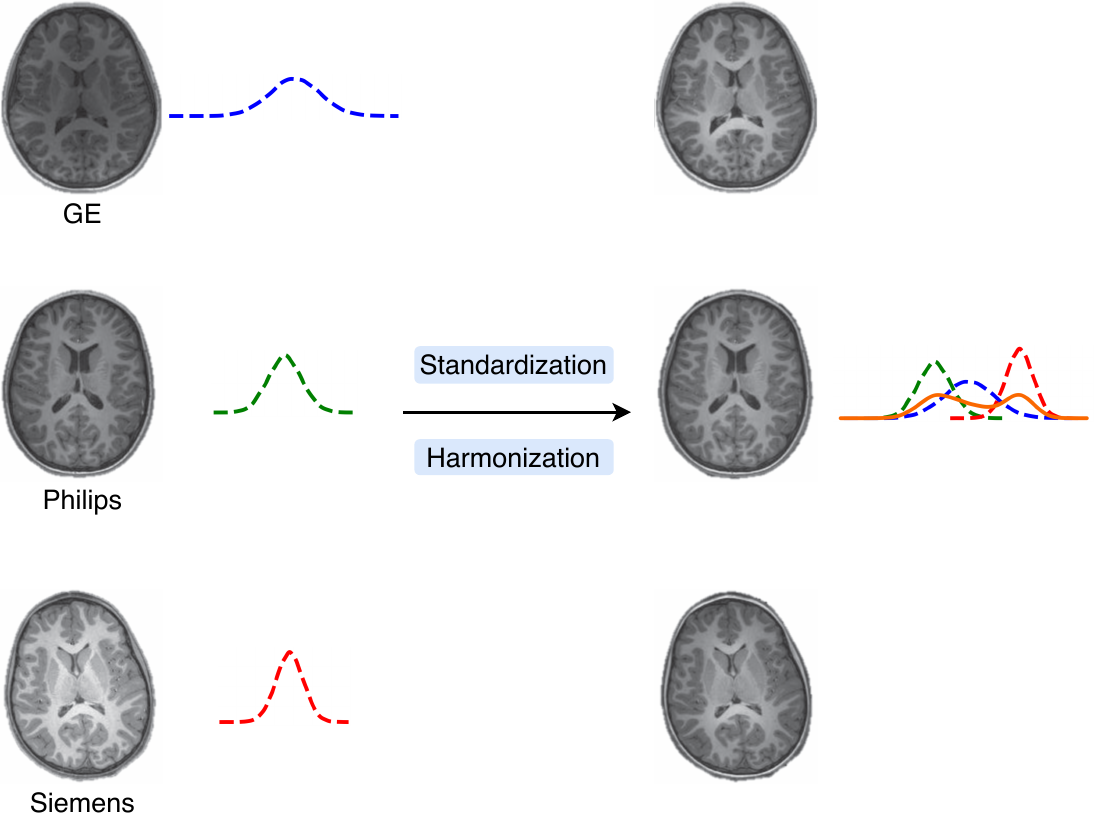}
    \caption{Illustration of Data Standardization and Harmonization. Multi-site T1-weighted MRI images from different scanners (GE, Philips, Siemens) exhibit variability in intensity distributions. The standardization process adjusts individual distributions to a common scale, while harmonization ensures alignment across datasets for improved consistency~\citep{liu2024learning}.}
    \label{fig:fl_harmonization}
\end{figure}

Healthcare data are often collected and stored in diverse and proprietary formats that do not always adhere to international standards and terminologies, complicating data linkage and reuse. 
For example, structured clinical data usually contains features that vary with differences in clinical practice across institutions~\citep{petersmann2019definition}, such as diabetes diagnosis, which can involve different glucose measurement methods with varying cut-off points, resulting in hidden heterogeneity that may be overlooked in subsequent statistical analyses.
Additionally, language differences across institutions, especially in multilingual regions like the European Union, pose additional challenges in standardizing and harmonizing data. Medical terminology and clinical reports may be documented in different languages, complicating data interpretation and analysis across borders.
A crucial step before implementing FL in healthcare is to ensure data standardization, harmonization, and interoperability across different cohorts, which are key to the success of FL (Figure~\ref{fig:fl_harmonization}).
\par
Most simulation studies processed data centrally and generate artificially partitioned datasets without considering the distributed nature of various data silos. This oversight extends to the lack of discussion on how datasets at each client are curated for use in experiments. 
Despite this, almost all FL frameworks assume the input data is preformatted for model training or preprocessing pipelines. This assumption leads to significant frustration and delays, as the burden of data export and conversion typically falls on clinical data managers who may lack the necessary budget and training.
Moreover, among the included studies, only two performed quality or integrity checks on the data. \cite{gad2022federated} excluded samples with impossible values (e.g., negative heart rates) and inconsistent feature values, while \cite{shaik2022fedstack} used Principal Component Analysis to filter out noise. 
Few studies addressed structural or informative missingness, with only \cite{alam2023fedsepsis} and \cite{sun2021fedio} considering imputation methods while also deleting features with high missingness rates. Poor quality imputation and handling of non-random missingness can bias model training. 
Additionally, no studies considered language differences in medical terminology and clinical reports across borders.

\paragraph{\textbf{Recommendations \& Opportunities}}
\begin{itemize}
    \item \textit{Adopt Standardized Formats:} Data standardization, harmonization, and interoperability across clients can be facilitated through formats such as Fast Healthcare Interoperability Resources (FHIR) for electronic health records (EHR) data and Digital Imaging and Communications in Medicine (DICOM) for imaging data.
    \item \textit{Simplify Data Preparation:} FL frameworks targeting healthcare research must simplify data preparation and ensure interoperability with standard data formats. This approach eases the burden on clinical data managers and improves data reusability.
    \item \textit{Use Consistent Protocols:} Collaborating clients must use consistent preprocessing protocols to standardize data to a Common Data Model (CDM) such as the Observational Medical Outcomes Partnership (OMOP)~\citep{lee2023feasibility}. Harmonizing healthcare data to a CDM like OMOP ensures it is interoperable with other clinical datasets, enabling effective merging and analysis across distributed sources and platforms~\citep{mateus2024data}.
    \item \textit{Automate Data Pipelines:} Extraction, transformation, and loading pipelines that automate the conversion of raw data to analysis-ready/training-ready data are needed to further simplify data standardization and harmonization. 
    \item \textit{Ensure Secure Access:} Secure access to fully standardized, harmonized, and interoperable large datasets through encryption methods can significantly accelerate clinical research within the federation.
    \item \textit{Address Language Differences:} In multilingual regions like the European Union, language differences in medical terminology can hinder data interpretation. Translation and normalization techniques, along with Large Language Models (LLMs), can assist in automatic translation and ensuring consistent terminology, thereby improving data interoperability for FL applications.
\end{itemize}

\subsubsection{Issues in Data Partition}
Among all included studies, only a few leveraged natural splits to replicate data collection processes across different hospitals or institutions. For instance, \cite{chakravarty2021federated} employed the CheXpert dataset, \cite{kaissis2021end} worked with the chest X-ray dataset and \cite{lu2022federated} extracted metadata from Tissue Source sites in the TCGA dataset for their studies.
These datasets naturally reflect the heterogeneity found in real-world clinical data across hospitals and institutions, making them more suitable for FL studies in healthcare~\citep{andreux2020siloed,baheti2020federated}.
\par
Simulation studies typically used heuristics to artificially create heterogeneous data partitions from a pooled dataset, assigning these partitions to simulated clients, as illustrated in Figure~\ref{fig:fl_simulationVSdeploy}. 
Common synthetic partitioning methods for classification tasks include assigning samples from a limited number of classes to each client, using Dirichlet distribution sampling on class labels, and employing the Pachinko Allocation Method (PAM) when labels have a hierarchical structure~\citep{mcmahan2017communication}. For regression tasks, Gaussian Mixture clustering based on t-SNE feature representations has been used to partition datasets among clients~\citep{philippenko2020bidirectional}. 
\par
Nonetheless, synthetic partitioning methods may not accurately reflect the intricate heterogeneity found in real-world scenarios~\citep{andreux2020federated}.
Examples from digital histopathology illustrate the limitations of synthetic partitioning methods~\citep{chen2024think}. In digital histopathology, tissue samples are extracted, stained, and digitized, leading to data heterogeneity due to factors such as patient demographics, staining techniques, physical slide storage methods, and digitization processes. Although advancements in staining normalization have reduced some heterogeneity, other sources remain challenging to replicate synthetically, and some may even be unknown~\citep{howard2021impact}. These underscore the necessity of conducting cohort experiments with natural splits to ensure that FL models are robust across varied clinical settings. This issue also extends to other areas, including radiology, dermatology, and retinal image analysis.
\par
Even among studies that adopted synthetic partitioning methods, the strategies employed are often limited, primarily focusing on scenarios such as quantity skew. These studies addressed only a narrow aspect of heterogeneity. For instance, label skew, where the distribution of labels differs across clients, and feature skew, where clients have different feature distributions, are frequently overlooked. As a result, the synthetic partitions created in these studies may not adequately represent the complex and varied heterogeneous conditions, potentially leading to less robustness in diverse healthcare environments.
In Section~\ref{skew}, we provide a comprehensive discussion of various types of skew and heterogeneity.
\par
Another significant issue is the lack of clear definitions and descriptions for train and test set partitions across clients in many studies. 
Among the studies included in this review, 84\% did not explicitly define how these partitions are handled for each client, leading to potential ambiguity in evaluating model performance. This concern is particularly critical in the context of personalized FL, where each client’s test set should be unique to accurately reflect individual data distributions. 

\paragraph{\textbf{Recommendations \& Opportunities}}
\begin{itemize}
    \item \textit{Complement Simulation Studies with Real-World Data Evaluations:} While simulation studies using artificially partitioned datasets can provide valuable insights, it is essential to validate these findings through evaluations on real-world, naturally partitioned datasets. This multi-stage evaluation process ensures that models are tested in both controlled environments and realistic deployment scenarios, improving their generalizability and robustness~\citep{kairouz2021advances}.
    \item \textit{Adaptive Partitioning Based on Data Distribution:} Researchers should consider using adaptive partitioning techniques that account for the underlying data distribution and specific characteristics of each client’s data. This can create more realistic and representative partitions, especially in scenarios where data is highly heterogeneous.
    \item \textit{Incorporate Multiple Types of Skew:}  Researchers should broaden the scope of their synthetic partitioning strategies to include not just quantity skew but also label skew and feature skew. This would create a more realistic representation of the heterogeneity found in real-world datasets, allowing FL models to be more robust and generalizable across diverse clinical settings~\citep{li2022federated}.
    \item \textit{Retaining Client-Specific Test Sets:} Consider the non-IID nature of data in FL, we suggest that researchers retain a portion of data within each client as a dedicated test set rather than relying on a single, global test set for all clients. This approach provides a more accurate and reliable evaluation, reflecting the unique data distributions of each client, which is particularly important in personalized FL scenarios~\citep{kairouz2021advances}.
    \item \textit{Incorporation of Domain Knowledge:} Incorporate domain knowledge into the partitioning process can enhance the relevance of synthetic data splits. In medical imaging, for example, understanding the clinical context and variability in imaging protocols across different institutions can inform more meaningful data partitioning strategies.
    \item \textit{Transparency and Reproducibility:} Researchers should provide detailed documentation of their data partitioning strategies, including the rationale behind their choices and any domain-specific considerations. This transparency will enable others to replicate and build upon their work effectively.
\end{itemize}

\subsection{Model}
\subsubsection{Inadequate Model Selection and Development}
The studies reviewed exhibit a wide range of model complexities, from advanced, parameter-heavy architectures to traditional ML techniques.
However, several issues persist in model selection and development.
\textit{Firstly}, there is an over-reliance on complex models, particularly CNNs, which dominate due to their high performance but pose challenges in resource-constrained healthcare environments. Their complexity complicates reproducibility and generalizability across different settings, particularly when custom architectures are involved.
\textit{Secondly}, the lack of standardization in model selection leads to variability in methodologies, making it difficult to compare results across studies and generalize findings. This inconsistency hampers the ability to benchmark performance across different FL applications.
\textit{Moreover}, simpler and more interpretable models are underutilized. While deep learning models offer high performance, traditional ML algorithms, which are easier to interpret and less resource-intensive, are often overlooked. These models could be more suitable for certain healthcare applications where interpretability and transparency are critical for clinical decision-making, offering a practical alternative that balances performance with the need for clarity and trustworthiness in healthcare settings.
\textit{Another challenge} is the limited focus on personalization. Many studies prioritized a single global model, which may not be optimal for all clients due to the heterogeneity in healthcare data. Personalized FL approaches, tailored to individual client data distributions, remain underdeveloped and require further research.
\textit{Lastly}, scalability concerns arise with complex models, particularly in large-scale healthcare networks. The communication and computational overhead of training such models can become prohibitive, highlighting the need for more scalable FL solutions to ensure practical deployment.

\paragraph{\textbf{Recommendations \& Opportunities}}
\begin{itemize}
    \item \textit{Standardization for Research and Benchmarking:} Standardizing the process of model selection and development is essential for FL research, particularly in benchmarking and comparative studies. This standardization does not aim to limit innovation but to provide a consistent framework for evaluating models across diverse FL applications an settings. For instance, establishing shared protocols for selecting baseline models, defining performance metrics, and validating results can significantly enhance reproducibility and comparability. Such a framework encourages innovations that are both rigorous and generalizable, enabling the development of practical solutions tailored to the unique challenges of FL, such as data heterogeneity and resource constraints~\citep{ogier2022flamby}.
    \item \textit{Exploring Simpler, More Interpretable Models:} Researchers should not overlook simpler, traditional ML models, particularly in scenarios where interpretability is crucial. These models can be more practical and equally effective in certain healthcare applications, providing a balance between performance and interpretability~\citep{argente2024interpretable,li2023towards}. Researchers should also consider developing simpler and lightweight models that can be deployed in resource-constrained environments without sacrificing performance.
    \item \textit{Enhancing Personalization in FL:} More work is needed to develop and refine personalized FL approaches that can adapt to the diverse data distributions encountered in healthcare. This could involve hybrid models that combine the strengths of both global and local models, as well as more sophisticated techniques for model adaptation~\citep{zhang2023grace}.
    \item \textit{Addressing Model Heterogeneity:} Given the diverse requirements and constraints across different institutions, it is crucial to develop FL strategies that can effectively manage model heterogeneity. This includes exploring federated ensemble learning methods, which allow the aggregation of heterogeneous models and can lead to more robust and accurate predictions~\citep{mabrouk2023ensemble}.
    \item \textit{Improving Scalability and Efficiency:} Future studies should prioritize the design of scalable FL models that can handle an increasing number of clients without excessive computational and communication costs. This could involve the development of more efficient algorithms and the use of federated distillation techniques to reduce model size and complexity~\citep{ullah2023scalable}.
\end{itemize}

\subsubsection{Negligence in Initialization}
Most of the included studies began federated training from a random initialization, a method that, while effective in IID scenarios, can be less optimal for handling non-IID data. In healthcare, where data distribution often varies significantly across institutions due to differences in patient demographics, clinical practices, or data collection methods, random initialization can lead to slower convergence, increased communication costs, and potentially suboptimal local optima~\citep{nguyen2023where,chen2023importance}.
\par
A significant issue is the lack of standardization in model initialization approaches. Many studies either adopted random initialization without justification or entirely omitted the description of their initialization method. This inconsistency can result in significant variations in model performance and convergence rates, making it difficult to compare results across different studies and settings. Additionally, if the initial model is biased towards the data distribution of certain participants, it might not perform well across all clients, leading to fairness issues and suboptimal overall performance.
Moreover, these challenges are further exacerbated by the heterogeneity of computational resources available across institutions. Some advanced initialization methods, such as those involving foundation models or pretraining on large-scale datasets, may be computationally expensive and thus infeasible for resource-constrained participants~\citep{nguyen2023where,li2024begin,peng2024depth}.
\par
Personalization in model initialization is another underexplored area. Personalized initialization techniques, which tailor the starting point to the specific data distribution of each client, are critical for improving local model performance and accelerating convergence. However, research into these techniques, such as model-agnostic meta-learning and partial initialization for finding a good initialization, remains limited within FL for healthcare~\citep{fallah2020personalized,sun2021partialfed}.

\paragraph{\textbf{Recommendations \& Opportunities}}
\begin{itemize}
    \item \textit{Transfer Learning for Initialization:} Utilize pretraining or foundation models for initialization can provide a strong starting point for FL. This approach has been shown to not only speed up convergence but also mitigate the effects of both data and system heterogeneity~\citep{li2024begin,peng2024depth,nguyen2023where}, potentially closing the performance gap between FL and centralized learning~\citep{chen2023importance}. However, researchers should carefully consider the similarity between the source and target datasets to avoid negative transfer effects.
    \item \textit{Personalized Initialization:} Implement personalized initialization methods, such as model-agnostic meta-learning and partial initialization, can help customize the starting point of the model based on local data characteristics, further enhancing model performance on individual clients and improving overall system convergence~\citep{fallah2020personalized,sun2021partialfed}.
    \item \textit{Standardization of Initialization Procedures:} Standardizing initialization in FL is essential for ensuring consistency, reproducibility, and comparability across studies. It provides a common baseline for benchmarking and reduces variability in outcomes~\citep{chen2023importance}. For example, Nguyen et~al.~\citep{nguyen2023where} showed that consistent initialization improves convergence rates, especially under non-IID data, while pretraining-based methods help address system heterogeneity~\citep{li2024begin}. Such standardization does not hinder innovation but establishes a foundation for exploring advanced techniques like meta-learning and hybrid initialization, ensuring their broader applicability in diverse healthcare settings.
    \item \textit{Consideration of Resource Constraints:} When selecting initialization methods, researchers should account for the varying computational resources across participants. Techniques that balance initialization quality with computational feasibility, such as hierarchical model training or using lightweight pretrained models, are critical to ensuring broader applicability of FL in healthcare~\citep{li2024begin,peng2024depth,nguyen2023where}.
\end{itemize}

\begin{figure}[t]
    \centering
    \includegraphics[width=0.6\columnwidth]{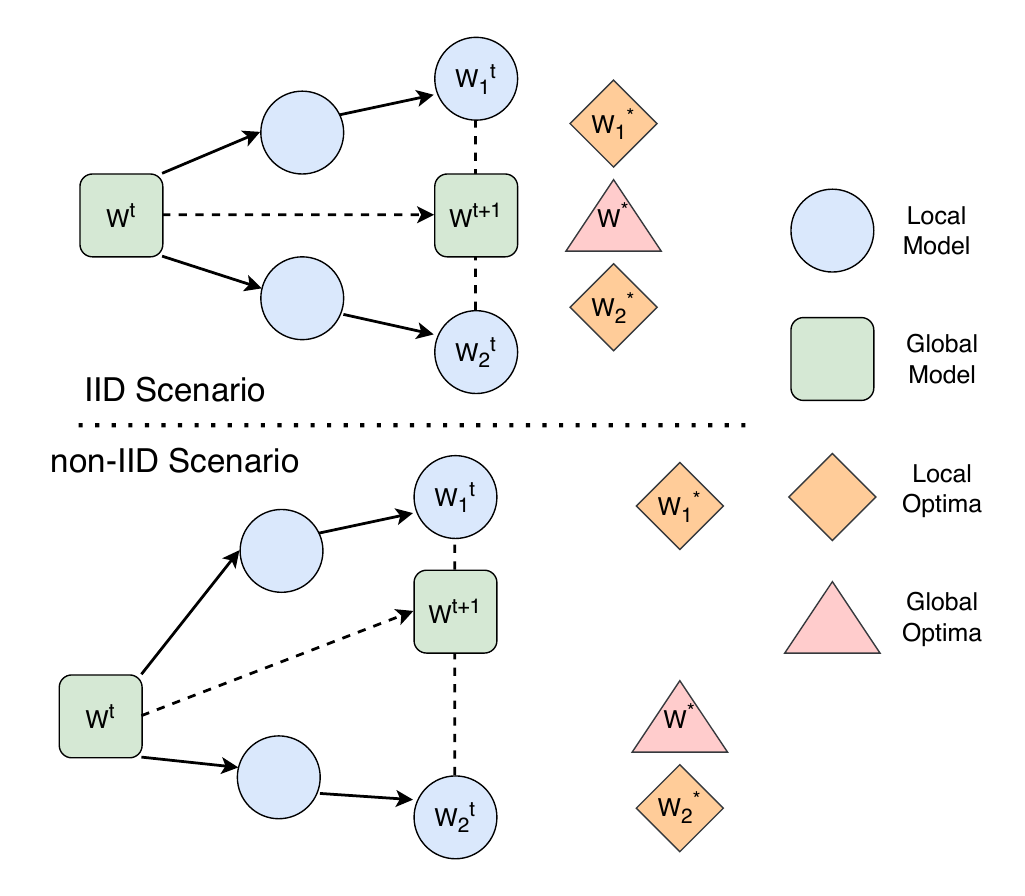}
    \caption{Drift issue in non-IID. In IID scenario, the global optima $w^*$ aligns closely with the local optima $w^*_1$ and $w^*_2$. Consequently, the aggregated model $w^{t+1}$ remains near the global optima. However, in non-IID scenario, the global optima $w^*$ may be significantly distant from $w^*_1$, leading to $w^{t+1}$ being far from the ideal solution.}
    \label{fig:fl_noniidVSiid}
\end{figure}

\subsection{Optimization}
\subsubsection{Heterogeneity Issues}\label{skew}
In FL for healthcare, heterogeneity refers to the variability in data, models, and systems across different hospitals and institutions. This variability poses significant challenges to FL’s performance and its ability to generalize well across diverse environments. The key types of heterogeneity in FL for healthcare include \textit{statistical heterogeneity}, \textit{model heterogeneity}, and \textit{system heterogeneity}.

\begin{figure*}[t]
    \centering
    \includegraphics[width=.79\linewidth]{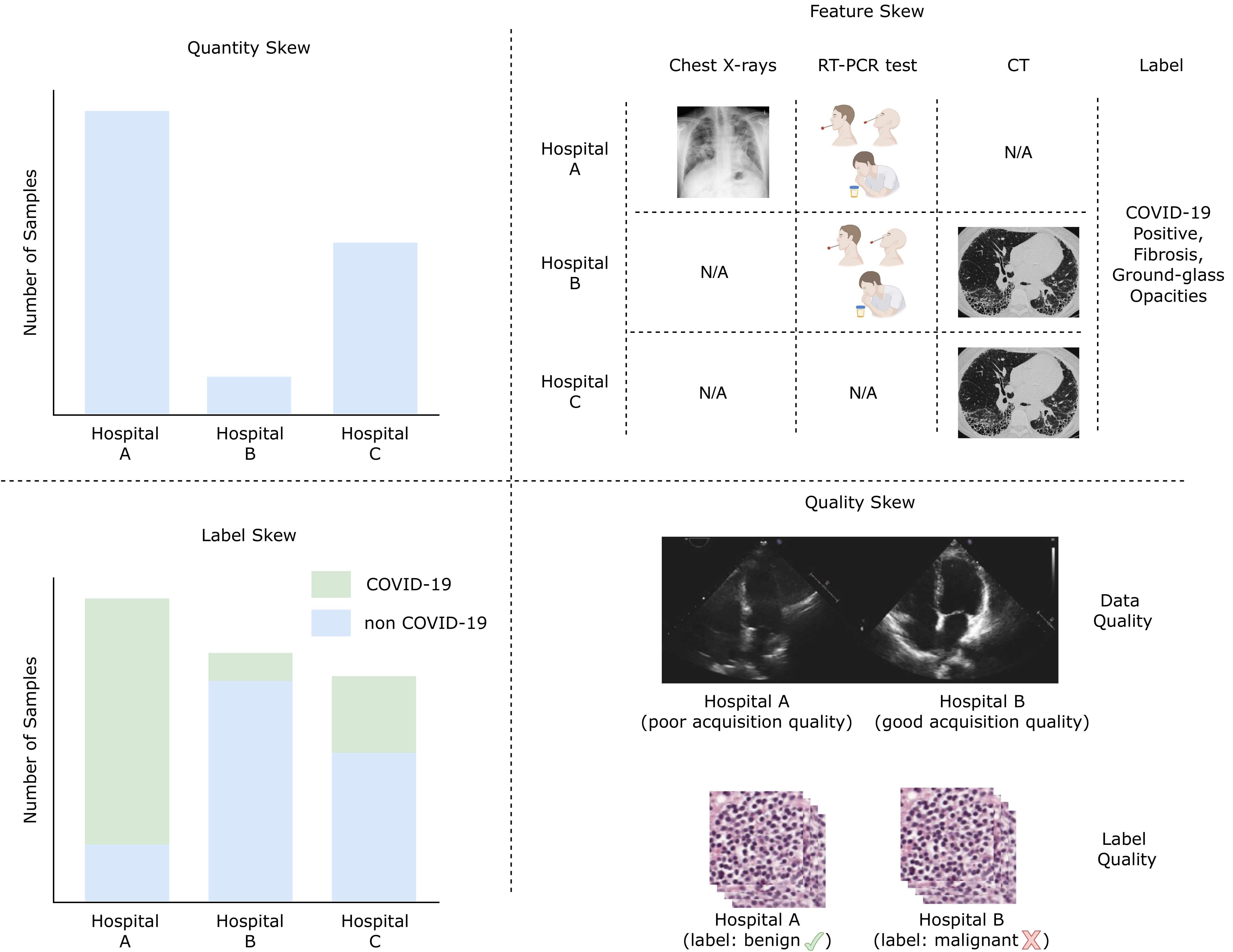}
    \caption{Illustration of different skew forms in statistical heterogeneity, including quantity, feature, label, and quality skew. Medical image sources: Chest X-rays~\citep{nih_chest_xrays}, CT scans~\citep{radiopaedia_CT}, Ultrasound images~\citep{creatis_camus_ultrasound}, and Whole Slide Imaging (WSI)~\citep{camelyon17_pathology}.}
    \label{fig:fl_skew}
\end{figure*}

\textit{Statistical heterogeneity} arises due to the non-IID nature of healthcare data across various institutions, which is characterized by demographic differences, instrumentation biases, distinct data acquisition protocols, and human operations, etc~\citep{guan2021domain}. 
For instance, variations in CT scan quality across sites can lead to inconsistencies in the correlation between imaging data and corresponding site-specific EHR data. These inconsistencies severely degrade FL performance, with accuracy drops of up to 50\%~\citep{mcmahan2017communication}, necessitating additional communication rounds for convergence~\citep{chen2021asynchronous}.
Statistical heterogeneity can also result in clients overfitting to their local data, leading to poor generalization on data from other clients, making simple parameter averaging an ineffective aggregation strategy~\citep{mora2024enhancing}.
Since local models are optimized towards different local optima, the aggregated global model may drift from the true global optima, causing a biased minimum and significantly slowing down convergence as illustrated in Figure~\ref{fig:fl_noniidVSiid}.
In healthcare, statistical heterogeneity can be broadly characterized by four forms, including:
\begin{itemize}
    \item \textit{Quantity Skew.} The number of training samples differs greatly across clients, leading to imbalanced data distributions. Models tend to optimizing for clients with more data, potentially neglecting those with less, further reducing the generalization ability~\citep{li2022federated}.
    \item \textit{Label Skew.} The distribution of labels varies across clients. For instance, in the context of COVID-19, hospitals in regions heavily impacted by the pandemic may have a higher proportion of positive cases, whereas other regions might have predominantly negative cases. This label imbalance can lead to biased models~\citep{ghassemi2020review}. 
    \item \textit{Feature Skew.} Different clients may have access to different features for the same sample cohort. For example, some institutions might only have access to EHR data, while others may have additional imaging modalities like X-rays or MRIs~\citep{fallah2020personalized,kline2022multimodal}. This is especially common in VFL scenarios.
    \item \textit{Quality Skew.} This arises from varying data quality across clients, often due to issues like label noise, data acquisition noise, or processing discrepancies. Clients with high-quality, accurately labeled data contribute more effectively to model learning, while those with noisy or inaccurate labels can introduce errors, undermining the model's generalization and convergence~\citep{wang2024federated}. 
\end{itemize}
Figure~\ref{fig:fl_skew} provides an illustration of these skew forms.
Numerous studies have been focusing on one of these skews in healthcare. 
\cite{qu2022handling} introduced a generative replay method by employing a VAE to synthesize medical images, enabling clients to train on a combined dataset of real and generated data, thus mitigating data quantity skew.
\cite{jiang2022harmofl} employed a frequency-based approach for medical data harmonization in FL, where images are processed in the frequency domain to retain local phase information and synchronize amplitudes across clients, thereby mitigating feature skew.
\cite{chen2024think} leveraged a Dirichlet distribution for modeling categorical probabilities in medical data, applying uncertainty calibration and diversity relaxation to enhance label annotation by focusing on samples with high uncertainty and low similarity, thus reducing label quality skew.
\par

\textit{Model heterogeneity} occurs when different clients use varying model architectures due to differences in hardware capabilities, data types, or specific institutional requirements. For example, one hospital may use a complex deep learning model for image analysis, while another uses a simpler model due to computational constraints. This divergence complicates the process of aggregating model updates in FL, as different architectures may have different fitting capabilities, performance characteristics, and requirements for convergence~\citep{lin2020ensemble}.
To address model heterogeneity, approaches like knowledge distillation have been employed, where a ``student" model is used to transfer knowledge from various ``teacher" models in different clients~\citep{lin2020ensemble}. However, these methods often require auxiliary public datasets for transferring knowledge, raising privacy concerns and increasing the computational burden on resource-constrained institutions~\citep{li2019fedmd,lin2020ensemble}.
An alternative approach is the use of a lightweight ``messenger" model, as proposed by \cite{xie2024mhpflid}, which carries concentrated information from one client to another, reducing the need for a full model exchange. This method allows for efficient aggregation and distribution of knowledge without the overhead of auxiliary public datasets.
\par

\textit{System heterogeneity} refers to the differences in computational capabilities, network architectures, and resource availability across clients. For example, some hospitals may have advanced computing infrastructure, while others may have limited hardware resources. This variability may affect the efficiency of FL, particularly in aggregating models trained on non-IID data~\citep{li2022federated,mabrouk2023ensemble}. Most studies have focused on addressing statistical or model heterogeneity, but system heterogeneity is equally important. Differences in hardware resources can lead to discrepancies in how models perform and converge across clients. Furthermore, the need for additional local computation and storage resources, as required by methods like knowledge distillation, can be a burden for institutions with limited resources~\citep{li2023data,li2019fedmd}.

\paragraph{\textbf{Recommendations \& Opportunities}}
\begin{itemize}
    \item \textit{Data Harmonization:} Apply data harmonization techniques locally at each client site to minimize variability in data distributions and improve the consistency of FL models~\citep{jiang2022harmofl,yan2020variation}.
    \item \textit{Data Synthesis and Augmentation:} Use data synthesis and augmentation techniques to generate additional data for underrepresented classes in the local data. Techniques such as GAN, VAE, and diffusion models can be leveraged to create synthetic data that preserves privacy while boosting model generalization. 
    \item \textit{Feature Alignment and Data Imputation:} Use methods like feature alignment and data imputation during training to ensure models learn from consistent data across clients. For example, FedHealth~\citep{chen2020fedhealth} demonstrated the ability to infer missing modalities in healthcare data through FTL.
    \item \textit{Bias Checks and Corrections:} Implement continuous monitoring for bias within and across clients throughout the FL process. Correct identified biases to prevent model performance degradation. For instance, \cite{chen2024think} utilized uncertainty calibration and diversity relaxation to dynamically correct annotations for high-uncertainty, low-similarity samples. 
    \item \textit{Advanced Optimization Techniques:} Enhance FL robustness against client drift and heterogeneity by employing advanced optimization techniques such as FedProx~\citep{li2020federated}, SCAFFOLD~\citep{karimireddy2020scaffold}, and MOON~\citep{li2021model}.
    \item \textit{Disentangled Representation Learning (RDL):} Integrate DRL into federated models to disentangle underlying invariant factors, making models more robust to data heterogeneity across different institutions~\citep{bercea2021feddis, luo2022disentangled}.
    \item \textit{Fine-Tuning and Personalization:} Improve model performance on domain-specific tasks by fine-tuning models locally with client-specific data and annotations. Extend techniques from meta-learning and multi-task learning to support personalized or device-specific modeling in FL~\citep{fallah2020personalized}.
    \item \textit{Comprehensive Heterogeneity Handling:} Address the complex and multifaceted nature of data skew and heterogeneity more comprehensively. Real-world scenarios often involve intricate combinations of quantity, label, feature, and quality skew, along with system and model heterogeneity. Developing more robust and flexible methods to handle these diverse challenges will be crucial for improving FL’s effectiveness in healthcare.
    \item \textit{Adaptive FL Frameworks:} Future research should focus on developing more adaptive FL frameworks capable of accommodating a wide range of model architectures and system configurations. Examples of adaptive frameworks include FedAdapt~\citep{wu2022fedadapt}, which adjusted client participation and resource allocation based on system heterogeneity, and AutoFL~\citep{kim2021autofl}, which used AutoML techniques to dynamically configure FL processes. Adaptive frameworks can also be realized through dynamic client selection and hierarchical aggregation (e.g., Clustered FL~\citep{sattler2020clustered}) to balance data quality, system resources, and communication efficiency.
\end{itemize}

\subsubsection{Open Domain Problem}
A key challenge in healthcare FL is the poor generalization of models to open domains, where unseen data lies beyond the federation's scope. A mere 14\% of the included studies have validated their methods on such external data, underscoring a significant research gap. Current FL strategies predominantly focus on boosting performance within the federation, frequently overlooking the essential need for model adaptability to new, unseen environments.
\par
Studies have shown that even slight differences in devices or acquisition protocols can result in a significant distribution shift, thereby reducing the model’s effectiveness when applied to new, unseen datasets. This issue is particularly acute in healthcare applications like diabetic retinopathy screening in fundus images, where the diversity of cameras and settings across different institutions can lead to poor model performance on external data. \cite{liu2021feddg} addressed this challenge by introducing a frequency-based domain generalization approach in FL. They enabled privacy preserving exchange of distribution information across clients through continuous frequency space interpolation and designed a boundary-oriented episodic learning scheme to expose local training to domain shifts and enhance model generalizability in ambiguous boundary regions.
However, the proposed method can be impractical for real-world applications due to its reliance on extensive network bandwidth and computational resources required for Fourier transform computation in frequency space interpolation.

\paragraph{\textbf{Recommendations \& Opportunities}}
\begin{itemize}
    \item \textit{Domain Generalization Techniques:} These techniques aim to create models that are robust to distribution shifts by learning domain-invariant features. Methods such as data augmentation~\citep{zhou2024federatedaug}, which generates diverse augmented features of client data to improve model robustness, adversarial training~\citep{zhang2023federateddg}, where domain-invariant representations are learned by minimizing domain discrimination, and meta-learning~\citep{liu2023domain},  which optimizes for rapid adaptation to new domains, have been explored to improve the generalization capabilities of FL models to unseen domains. 
    \item \textit{Transfer Learning and Fine-Tuning:} Apply transfer learning and fine-tuning on small amounts of unlabeled data from the open domain can help adapt the federated model to new environments~\citep{chen2020fedhealth}. Self-supervised learning techniques like contrastive learning can further facilitate this process~\citep{zou2023self}.
    \item \textit{Ensemble Learning:} Combine multiple models trained on different clients via sophisticated ensemble methods can enhance the robustness of the final prediction, making it more generalizable to open domains~\citep{lin2020ensemble}. 
    \item \textit{Data Synthesis and Simulation:} Generate synthetic data that mimics the characteristics of potential open domains can be used to pretrain or fine-tune the federated model, improving its generalization to unseen environments~\citep{liu2021feddg}.
\end{itemize}

\subsubsection{Burdens in Communication}
Communication is a significant bottleneck in the implementation of FL in healthcare. In FL, each client needs to frequently communicate with the central server. This communication can be orders of magnitude slower than local computation due to constraints such as bandwidth, latency, and power~\citep{ma2021communication}.
\par
The communication bottleneck in FL arises from several factors. 
\textit{First}, the number of clients involved in an FL system can be very large (e.g., wearables and IoMT), leading to significant communication overhead. Each communication round requires sending model updates between the clients and the central server, which can be expensive in terms of time and resources~\citep{ma2021communication,mcmahan2017communication}.
\textit{Second}, ensuring data privacy and security in FL is crucial, especially in sensitive domains like healthcare. The need for encryption and secure communication protocols adds to the computational and communication overhead. Encrypting model updates can significantly increase the size of the data being transmitted, further straining the communication channels and requiring more sophisticated algorithms to balance privacy and efficiency~\citep{mcmahan2017communication}.
\textit{Third}, as FL tasks become more complex, the size of the models involved increases. Modern large-scale models, such as large language models (LLMs) and foundation models (FMs), can have billions of parameters, resulting in model sizes that require significant bandwidth to transmit. This issue is exacerbated when using standard communication protocols like gRPC, which have size limits on single messages (e.g., 2 GB). Typical LLMs and FMs can exceed these limits, necessitating the model to be split into smaller chunks for transmission, adding additional overhead and complexity to the system~\citep{tayebi2023private}.
\par
Included studies primarily concentrated on enhancing communication encryption techniques, with the aim of either reducing the volume of data exchanged~\citep{lian2022deep,brisimi2018federated} or minimizing the number of communication rounds~\citep{souza2021multi}.
Additionally, several studies explored methods for achieving fully decentralized communication without central server~\citep{tedeschini2022decentralized,lu2020decentralized}.
Improvements in encryption often involved methods for securely sharing secret keys among clients~\citep{sav2022privacy,chen2020achieving}, encryption mechanisms for safeguarding exchanged data~\citep{li2021federated,zhang2022homomorphic}, and techniques for perturbing model outputs at each client using a secret key~\citep{park2022multi}. 
To decrease the amount of data transmitted, some studies proposed transferring only a subset of model parameters~\citep{sun2021partialfed,lu2022personalized,ma2021communication} or employing strategies like compressing~\citep{wang2022privacy}, masking~\citep{kerkouche2021privacy}, and quantizing gradients~\citep{ma2021communication} or model outputs before exchange~\citep{li2023data}. 
Reducing the number of communication rounds was addressed through model design~\citep{tong2022distributed,thakur2021dynamic,qu2022handling}, aggregating updates based on elapsed time instead of epochs~\citep{wang2022privacy,maj2021communication}, and evaluating the potential benefit of an update before communication~\citep{chen2020achieving}. 
Other studies focused on detecting attacks during communication~\citep{cholakoska2021differentially}, developing authentication systems for clients~\citep{wang2023privacy}, and improving client management systems~\citep{li2022contract,zhang2022homomorphic}.
\par

\paragraph{\textbf{Recommendations \& Opportunities}}
\begin{itemize}
    \item \textit{Model Optimization:} Utilize parameter-efficient, lightweight models such as MobileNet and SqueezeNet to reduce computation and memory overhead. Explore model pruning~\citep{wen2016learning} and low-rank adaptation (LoRA)~\citep{hu2021lora} techniques to decrease model size or regularize model weights without significantly impacting accuracy. Consider split learning~\citep{gawali2021comparison, thapa2022splitfed} to share intermediate activations or a subset of the model instead of the full model, thus reducing communication costs.
    \item \textit{Gradient Compression:} Implement gradient compression methods, including sparsification and quantization, to reduce the size of model updates~\citep{rothchild2020fetchsgd}. Integrate AI-driven compression techniques to dynamically adjust quantization and sparsification levels based on model performance and network conditions.
    \item \textit{Reduce Communication Rounds:} Minimize the number of communication rounds by adopting methods such as One-Shot FL~\citep{guha2019one}, which requires only a single round of communication to achieve effective model training. 
    \item \textit{Adopt Decentralized Approaches:} Explore decentralized training methods to eliminate the need for a central server, thereby alleviating communication bottlenecks and improving scalability.
    \item \textit{Knowledge Distillation:} Use knowledge distillation techniques to transmit only essential distilled knowledge, such as logits from the final layer, rather than the entire model~\citep{li2019fedmd, guha2019one,madni2023federated}. This approach can reduce communication overhead while maintaining model performance.
    \item \textit{Data Distillation:} 
    Apply data distillation to generate synthetic data summaries or distilled representations of the original datasets or updates. Clients can then share smaller, more concise updates with the central server, enhancing communication efficiency~\citep{xiong2022feddm, goetz2020federated}.
    \item \textit{Develop Specialized Protocols:} 
    Design and implement lightweight communication protocols tailored for FL, focusing on bandwidth efficiency and robustness against network instability.
    \item \textit{Design Scalable and Flexible Framework:} Develop FL frameworks that adapt to various communication environments and model complexities. Incorporate dynamic communication schedules to optimize efficiency and performance.
\end{itemize}

\subsubsection{Plain Convergence Analysis}
Federated optimization in healthcare aims to adapt models to local data distributions while integrating global information. 
The inherent heterogeneity across hospitals and institutions often leads to instability and slow convergence in federated training~\citep{li2020federated}. 
However, comprehensive convergence analysis is frequently lacking in current studies.
\par
Most studies tend to provide only plain convergence analysis, focusing on reporting metrics such as the number of local epochs, communication rounds, and overall convergence time. 
While these metrics are useful, they do not offer a deep theoretical understanding of the convergence dynamics. 
This lack of rigorous analysis limits our understanding of how and why certain FL algorithms perform well (or poorly) in specific healthcare applications.
\par
Only a handful of studies~\citep{lu2020decentralized, kerkouche2021privacy, ma2021communication, brisimi2018federated} have focused on the convergence of FL in healthcare settings. 
These studies typically relied on Stochastic Gradient Descent (SGD) as the foundational optimization method due to its effectiveness in smooth optimization problems, under assumptions such as the existence of lower bounds, Lipschitz smoothness, and bounded variance~\citep{lu2020decentralized,kerkouche2021privacy}. 
However, SGD-based FL algorithms often struggle with nonsmooth optimization problems, which are common in healthcare data due to irregularities in data distributions and the presence of outliers.

\paragraph{\textbf{Recommendations \& Opportunities}}
\begin{itemize}
    \item \textit{Theoretical Convergence Guarantees:} Establish rigorous theoretical methods that provide convergence guarantees for FL in healthcare. This includes deriving bounds on convergence rates and understanding the conditions under which proposed algorithms perform optimally, particularly in non-IID data settings. While complete boundary analyses in large-scale non-IID settings remain challenging due to data noise and complexity, partial modeling of noise, such as using bounded variance assumptions or noise-resilient gradient estimators, has shown promise in existing studies~\citep{li2020federated, mishra2022noise}. Future research could explore hybrid approaches that combine empirical noise estimation with theoretical bounds to enhance the practical relevance of convergence analysis.
    \item \textit{Real-World Validation:} Prioritize validating convergence analysis and algorithmic improvements on real-world healthcare datasets with natural split rather than relying solely on synthetic partitions. This will ensure that the proposed methods are practical and effective in real healthcare environments.
    \item \textit{Advanced Optimization:} Non-convexity, non-smoothness, and heterogeneity are not universal across all healthcare data. For instance, MRI images tend to be homogeneous, while blood test data exhibit more heterogeneity. A unified optimization method may not address all these challenges. Tailored optimization strategies, including smooth methods for homogeneous data and non-smooth, non-convex methods for heterogeneous data, should be explored. Hybrid approaches could also be considered to optimize convergence and stability in diverse healthcare applications.
    \item \textit{Principled Communication Termination:} Implement principled methods for terminating communication rounds, potentially based on the performance of the global model at each client on a validation cohort or evaluation data held at the central aggregator. Early stopping based on local convergence could also be beneficial, as it would reduce unnecessary computation and communication costs.
\end{itemize}

\subsubsection{Temporal Dynamics and Revoke Issues}
Healthcare data's inherent time dependence is critical, especially for diseases with distinct progression or treatment timelines, such as cancer and chronic conditions like diabetes. However, included studies often overlooked these temporal dynamics when partitioning data, potentially leading to models that inaccurately reflect disease progression. 
For example, COVID-19 characteristics, such as ground-glass opacities in lung CT scans, evolve with the disease's progression~\citep{Pan2020}. Ignoring such temporal dynamics can result in models that are overfitted to specific stages of a disease and unable to generalize across different phases~\citep{li2019recurrent,li2020mv,li2020unified}. 
Furthermore, the dynamic nature of data at each participating hospital or institution complicates the situation. Hospitals continuously acquire new data and may also remove or modify existing data due to errors or other factors.
This dynamic data environment requires FL models to adapt without frequent retraining, as new data might cause model drift, while data removal can leave critical gaps in the model’s understanding, particularly if the removed data represents rare or critical cases.
\par
Another critical but overlooked issue in FL is data revocation, which becomes necessary when specific data must be withdrawn due to privacy concerns, regulatory requirements, patient or participant requests, misdiagnosis, invalidated prior diagnoses, or medical misconduct. 
Current FL setups, designed for iterative data aggregation, struggle with ``unlearning'' specific contributions without requiring complete model retraining. 
Emerging research highlights the need for mechanisms that allow for efficient data revocation without compromising the integrity of the model. For instance, methods have been proposed to facilitate client ``unlearning'' in FL, enabling the removal of data contributions from specific clients without significant degradation in model performance~\citep{wang2024forget}. This is critical in healthcare, where patient consent may be withdrawn, new privacy regulations may require data deletion, or misdiagnosed and fraudulent data could undermine model integrity.

\paragraph{\textbf{Recommendations \& Opportunities}}
\begin{itemize}
    \item \textit{Incorporate Temporal Dynamics:} FL frameworks should be adapted to account for the temporal aspects of healthcare data. This could involve time-aware partitioning of data and the development of models that can learn from time-series data, ensuring better generalization across different stages of disease progression. 
    \item \textit{Support Dynamic Data Management:} FL models should include mechanisms for continuous learning to adapt to the addition and removal of data within the federation. Techniques like incremental learning or online learning could be employed to keep models updated with the latest data while minimizing the need for complete retraining.
    \item \textit{Data Revocation:} Develop and integrate efficient data revocation techniques in FL frameworks. Approaches such as machine unlearning~\citep{wang2024forget} can be refined to allow the removal of specific data contributions while minimizing the impact on overall model performance. This will be critical for maintaining compliance with privacy laws and upholding patient rights in healthcare applications.
    \item \textit{Active Learning Integration:} Incorporate active learning strategies into FL to selectively query the most informative data points during the training process. This would help in focusing on critical data that improves model performance over time, especially in cases where temporal dynamics are at play.
\end{itemize}

\subsubsection{Synchronization Issues}
Synchronization of updates across different clients also poses significant challenges for FL in healthcare.
The variation in computational resources, network conditions, and data availability among clients can lead to different training speeds and delayed model updates. 
\par
In \textit{Synchronous FL}, all clients must complete their training and send their updates before the global model aggregation, which is straightforward but can be inefficient. This approach works well in environments where data is immediately available, such as in a centralized hospital picture archiving and communication system (PACS).
However, in real-world scenarios, where data acquisition might be delayed due to network issues or the unavailability of input/output devices, synchronous updates can result in significant idle times and resource underutilization~\citep{sakib2021asynchronous}. Additionally, clients in an FL network, particularly smaller healthcare entities (e.g., wearables and IoMT devices), may not be active during every communication round, further delaying the global model update and potentially degrading overall system performance~\citep{mcmahan2017communication}.
\par
\textit{Asynchronous FL}, on the other hand, allows clients to send updates independently, without waiting for other clients to finish their training. This approach is more flexible and can accommodate variations in client availability and computational power, thereby improving the overall efficiency of the FL process. 
One study~\citep{sakib2021asynchronous} proposed an asynchronously updating FL architecture for cardiac activity monitoring and arrhythmia detection without the need for frequent synchronization. Another study~\citep{shiranthika2024adaptive} presented an adaptive asynchronous split FL scheme for medical image segmentation, which enhances training efficiency and model performance by allowing clients to operate at their own training speeds.
However, asynchronous updates can introduce challenges related to the consistency of the global model, as updates from slower or less reliable clients may arrive out of sync with the rest of the system.

\paragraph{\textbf{Recommendations \& Opportunities}}
\begin{itemize}
    \item \textit{Hybrid Synchronization:} Investigate hybrid synchronization methods that combine the benefits of synchronous and asynchronous updates. For example, employing synchronous updates for critical clients with high-quality data and asynchronous updates for less reliable or slower clients could balance consistency and efficiency.
    \item \textit{Consistency and Anomaly Check:} When employing asynchronous updates, it is crucial to assess the consistency and anomaly of updates received from different clients. Divergent updates, where one client's model update significantly differs from others, could indicate issues such as heterogeneity or anomalous behaviour in the training process. These discrepancies need to be carefully managed to prevent the global model from diverging.
    \item \textit{Asynchronous FL for Wearables and IoMT  Devices:} Asynchronous FL is particularly suitable for wearables and IoMT, where devices may have intermittent connectivity. Leveraging the Async-FL paradigm can pave the way for implementing the next generation of smart and remote healthcare monitoring systems at a mass scale.
\end{itemize}

\begin{table*}[t]
    \centering
    \caption{Summary of privacy and security attacks in FL applied to healthcare.}
    \resizebox{!}{50mm}{
    \begin{tabular}{c|m{60mm}|m{60mm}|m{16mm}|m{30mm}|m{37mm}|m{37mm}}
    \toprule
    Attacks & 
    \makecell[c]{Description} & 
    \makecell[c]{Risks} & 
    \makecell[c]{Attack\\Difficulty} &
    \makecell[c]{Impact\\Scope} & 
    \makecell[c]{Detection\\Methods} &
    \makecell[c]{Defense\\Strategies}\\
    \midrule
    \makecell[c]{Model Inversion\\ \citep{li2022e2egi}}
        & Reconstruct the actual samples (e.g., patient medical images or genetic data) from the model or updates
        & Potential leakage of original patient data
        & High 
        & Data privacy, patient confidentiality 
        & Anomaly detection in model outputs \& updates 
        & DP \\
    \hline
    \makecell[c]{Membership Inference\\ \citep{ijcai2020_431}}
        & Determining if a specific patient’s record was part of the model’s training set by analyzing its outputs
        & Compromises patient confidentiality, leading to potential unauthorized access to sensitive health information
        & Medium
        & Data privacy, patient confidentiality
        & Monitoring for unusual model behaviour, particularly overconfident predictions
        & Regularization techniques, DP, and robust model validation \\
    \hline
    \makecell[c]{Data Attribute Inference\\ \citep{melis2019exploiting}}
        & Infer individual sample's attributes (like race, gender, age) or gain aggregate statistical insights about the entire training set from model parameters and updates
        & Leakage of patient privacy
        & Medium
        & Data privacy, patient confidentiality, communication security
        & Gradient analysis, privacy audits
        & Secure aggregation, DP\\
    \hline
    \makecell[c]{Data Poisoning\\ \citep{nasr2019comprehensive}}
        & Malicious participants alter local data or labels to degrade the global model’s performance
        & Significant drop in model performance, potentially misleading patient diagnoses
        & Medium
        & Model accuracy, patient safety
        & Monitoring for abnormal model behaviour, statistical checks
        & Robust aggregation methods, anomaly detection in model updates \\
    \hline
    \makecell[c]{Model Poisoning\\ \citep{kalapaaking2023blockchain}} 
        & Malicious participants upload tampered model parameters to manipulate the global model
        & Global model may be intentionally corrupted, affecting its performance on real data
        & High
        & Model integrity, patient safety
        & Anomaly detection in model updates
        & Robust aggregation, DP \\
    \hline
    \makecell[c]{Denial of Service\\ \citep{salim2024articlesfederated}}
        & Disrupt the FL system by overwhelming it with requests or blocking normal data flow
        & Training process delays or interruptions, affecting time-sensitive medical applications
        & Medium
        & System availability and security, training efficiency
        & Monitoring network traffic for anomalies
        & Rate limiting, robust network design, redundancy mechanisms \\
    \hline
    \makecell[c]{HE Attack\\ \citep{zhang2022homomorphic}}
        & Decrypt encrypted model updates to access sensitive information
        & Encryption key leakage may lead to a breakdown in data protection, compromising patient privacy
        & High
        & Data privacy, model integrity
        & Encryption integrity checks, key management audits
        & Secure key management, SMPC \\
    \hline
    \makecell[c]{User Withdrawal Attack\\ \citep{tian2023robust}}
        & Participants withdraw from training, but their prior updates still affect the global model
        & Updates from withdrawn users may contain errors or malicious data, degrading the global model
        & Low
        & Model integrity
        & Tracking user participation and contribution consistency
        & Periodic re-evaluation of model contributions, machine unlearning\\
    \bottomrule
    \end{tabular}
    }
    \label{table:fl_risks}
\end{table*}

\subsection{Privacy and Security}
Two critical privacy and security issues exist in current studies. First, it is concerning that 62\% of the included studies did not encrypt model updates during communication. This lack of encryption leaves FL system vulnerable to interception, posing significant security risks. Second, statistical information such as sample sizes and distributions were often shared alongside model updates, particularly in FedAvg-based methods~\citep{adnan2022federated,ziller2021differentially}. This practice can expose participants with large datasets to targeted attacks if an adversary intercepts the communication or compromises the aggregator~\citep{tayebi2023private}.
\par
Despite the advantage of FL in healthcare without directly exchanging or sharing local data, it is not immune to privacy and security risks.
Adversaries can analyze changes in model updates over time to infer sensitive information or exploit system vulnerabilities to conduct targeted attacks using techniques like model inversion~\citep{wu2019p3sgd}, membership inference~\citep{li2021membership}, and poisoning~\citep{nasr2019comprehensive}.
Additionally, clients may unintentionally reveal private data during the FL process. This can happen when the client memorizes previous model and gradient updates, leading to the leakage of sensitive information~\citep{hatamizadeh2023gradient, fang2022dp}. Furthermore, methods involving the sharing of a few data samples for augmentation or disclosing local data distributions during knowledge transfer can also result in privacy breaches~\citep{xie2024mhpflid, madni2023federated}.
These adversarial and unintentional exposures undermine the privacy and security of the FL process, necessitating robust countermeasures and continuous vigilance to safeguard the integrity and confidentiality of the FL process.
\par
In Table~\ref{table:fl_risks}, we provide a comprehensive overview of various privacy and security attacks identified in the context of FL in healthcare.
By summarizing their key characteristics, specific risks they introduce, as well as potential defence strategies and other relevant factors. This table serves as a critical resource for understanding the complexities and vulnerabilities associated with FL in healthcare, offering insights into both the challenges and possible solutions for enhancing privacy and security in this domain.
\par
Among all included studies, methods for data privacy and security protection generally fall into two broad categories, namely \textit{cryptography} and \textit{perturbation} techniques. Each of these methods has its advantages and shortcomings, as summarised in Table~\ref{table:fl_risks_defence}.

\textit{Cryptography} encompass a variety of methods, with homomorphic encryption (HE)~\citep{liu2021machine} and secure multi-party computation (SMPC)~\citep{lindell2020secure} being among the most popular.
HE enables computations directly on encrypted data without the need for decryption. This ensures that data remains encrypted throughout processing, storage, and transmission, thus providing robust data security and privacy. HE allows operations on ciphertexts, with the results, once decrypted, accurately reflecting the outcomes of operations performed on the original plaintext data.
HE is particularly valuable in FL due to its ability to safeguard data privacy during computation. Recent research has highlighted HE’s effectiveness in various healthcare applications, including oncology and medical genetics. For instance, \cite{froelicher2021truly} demonstrated HE’s potential for truly private federated evaluations, and \cite{chen2020fedhealth} successfully utilized HE for model aggregation in FL.
While SMPC enables multiple parties to collaboratively compute a function over their combined data while keeping each party’s data private. Each participant holds a piece of encrypted or encoded data, and the computation is designed so that no party can access the others’ data or infer anything beyond the final result. SMPC ensures the confidentiality of both input data and computation results, making it robust against adversarial attacks and suitable for scenarios with multiple untrusted parties. It is increasingly used in healthcare and other sensitive applications to enhance privacy. For instance, research shows that SMPC can improve privacy in FL with medical datasets by addressing risks related to malicious models and enhancing the confidentiality of model aggregation~\citep{kalapaaking2022smpc, kalapaaking2023blockchain}.
\par
Nevertheless, HE typically involves high storage and computational overheads. Encrypted data requires significant processing power, and the complexity of HE schemes can introduce a single point of failure, where a single server manages all encrypted data~\citep{froelicher2021truly}. Additionally, managing encryption keys securely is crucial, as any compromise in key management can jeopardize the entire system’s security~\citep{huang2023differential}.
Also, SMPC often incurs high computational and communication overhead. The process of encrypting, encoding, and splitting data can be computationally intensive. Moreover, coordinating the computations across multiple parties requires substantial communication resources, which can become a bottleneck as the number of participants grows~\citep{huang2022robust}. 

%
\textit{Perturbation} techniques, with differential privacy (DP)~\citep{dwork2006differential} being the most prominent, are crucial for protecting sensitive healthcare data.
DP quantifies the risk of exposing individual data points and ensures that the inclusion or exclusion of any single data point has minimal impact on the model’s output by introducing randomness into the model’s results.
In FL, DP is typically implemented by adding noise to the local updates or gradients before they are aggregated. This noise is designed to mask the contributions of individual data points, thus making it difficult to infer any single data point from the model’s outputs. 
DP provides a quantifiable measure of privacy through parameters such as $\epsilon$ and $\delta$, allowing for a clear understanding of the trade-off between privacy and model utility.
Recent studies in healthcare have shown that FL with DP can achieve comparable accuracy to non-DP models in specific tasks, with a performance gap of less than 5\%, proving DP's efficacy with minimal accuracy compromise~\citep{adnan2022federated,ziller2021differentially,tayebi2023private}.
\par
However, the main challenge with DP is balancing the privacy budget with model performance. Adding noise typically reduces model accuracy by obscuring data patterns, which is a critical consideration in DP implementation~\citep{hatamizadeh2023gradient}. Managing the privacy budget $\epsilon$ involves a trade-off: a smaller $\epsilon$ enhances privacy but may reduce performance, and a larger $\epsilon$ offers less privacy protection. Effective management of this trade-off is essential, requiring careful attention to privacy needs and performance goals~\citep{kaissis2021end}.
Implementing DP also introduces additional computational overhead due to the noise addition and gradient clipping processes. This overhead can affect the scalability and efficiency of FL systems, particularly in scenarios with large numbers of clients or data~\citep{lee2021scaling}. 
\par

\begin{table*}[ht]
    \centering
    \caption{Comparison of Cryptography, perturbation, and other techniques used for privacy and security protection in FL for healthcare.}
    \resizebox{!}{53mm}{
    \begin{tabular}{c|c|m{70mm}|m{100mm}|m{100mm}}
    \toprule
    \makecell[c]{Category} & 
    \makecell[c]{Technique} & 
    \makecell[c]{Description} & 
    \makecell[c]{Advantages} & 
    \makecell[c]{Limitations}\\
    \midrule
    \multirow{2}{*}{Cryptography} & 
    \makecell[c]{Homomorphic\\Encryption} & 
    Enable computations on encrypted data without decryption, maintaining data encryption throughout & 
    \begin{itemize}
        \item Provide strong data security and privacy
        \item Useful for various healthcare applications
        \item Effective in federated evaluations and model aggregation
    \end{itemize} & 
    \begin{itemize}
        \item High computational and storage overhead
        \item Complex schemes may introduce single points of failure
        \item Key management challenges
    \end{itemize} \\
    \cline{2-5}
    & 
    \makecell[c]{Secure\\Multi-Party\\Computation} & 
    Allow multiple parties to collaboratively compute a function on their combined data while keeping data private & 
    \begin{itemize}
        \item Ensure confidentiality of input data and computation results
        \item Robust against adversarial attacks
        \item Enhance privacy in federated settings
    \end{itemize} & 
    \begin{itemize}
        \item High computational and communication overhead
        \item Scalability issues with increasing number of participants
        \item Coordination complexity
    \end{itemize} \\
    \hline
    \makecell[c]{Perturbation} & 
    \makecell[c]{Differential\\Privacy} & 
    Adds noise to local updates or gradients to mask individual data contributions, ensuring privacy & 
    \begin{itemize}
        \item Provide strong privacy guarantees with quantifiable privacy budget
        \item Achieve competitive performance in healthcare applications
    \end{itemize} & 
    \begin{itemize}
        \item Trade-off between privacy and model accuracy
        \item Additional computational overhead
    \end{itemize} \\
    \hline
    \multirow{2}{*}{Others} & 
    \makecell[c]{Blockchain} & 
    A decentralized ledger system that secures data sharing across all clients, reducing single points of failure & 
    \begin{itemize}
        \item Decentralized data sharing and management through distributed ledger
        \item Enhanced data integrity and security with no single point of failure
        \item Effective at preventing and mitigating poisoning attacks with verification schemes
    \end{itemize} & 
    \begin{itemize}
        \item High communication and computational costs, which may limit scalability
        \item Potential challenges in key management and the need for substantial processing power
    \end{itemize} \\
    \cline{2-5}
    & 
    \makecell[c]{Swarm\\Learning} & 
    A fully decentralized model training approach without a central aggregator & 
    \begin{itemize}
        \item Enhance resilience against attack
        \item Effective in handling non-IID data, making it suitable for diverse and heterogeneous datasets
        \item Dynamically integrate decentralized hardware infrastructures
    \end{itemize} & 
    \begin{itemize}
        \item Latency issues due to peer-to-peer communication, potentially slowing down the training process
        \item The absence of a central aggregator may limit certain coordination and optimization capabilities
    \end{itemize} \\
    \bottomrule
    \end{tabular}
    }
    \label{table:fl_risks_defence}
\end{table*}

\textit{Beyond} cryptography and perturbation methods, blockchain~\citep{kumar2021blockchain} and swarm learning (SL)~\citep{warnat2021swarm,saldanha2022swarm} have gained lots of attention for enhancing privacy and security in healthcare.
\textit{Blockchain} is increasingly being integrated with FL to address data security and trust issues by replacing the central server with a decentralized privacy protocol. The key advantage of blockchain lies in its distributed ledger system, which ensures that data is securely shared and maintained across all clients without relying on a central authority. This decentralization reduces the risk of single points of failure and enhances data integrity. Furthermore, blockchain's verification schemes play a crucial role in the FL process, helping to detect and mitigate threats such as poisoning attacks. For instance, a blockchain-based FL framework developed by \cite{chang2021blockchain} combined DP and gradient-verification protocols to enhance security in IoMT devices, significantly reducing the success rate of poisoning attacks in tasks like diabetes prediction. Another approach by \cite{rehman2022secure} utilized blockchain alongside an intrusion detection system to monitor and prevent malicious activities during model training, further securing federated healthcare networks. However, while blockchain offers robust security features, its integration with FL is often challenged by high communication and computational costs, which can limit its scalability and efficiency.
\textit{Swarm Learning} takes a different approach by decentralizing not just the privacy protocol but the entire model training process. In SL, there is no central aggregator, instead, decentralized hardware infrastructures work together to securely onboard clients and collaboratively generate a global model. This decentralized approach enhances the network’s resilience against attacks and is particularly effective in scenarios where data is non-IID, such as in the prediction of conditions like COVID-19 and leukemia~\citep{warnat2021swarm,saldanha2022swarm}. 
However, SL's reliance on peer-to-peer communication can introduce latency issues, which may slow down the training process, particularly in environments with varying network conditions.
\par

\paragraph{\textbf{Recommendations \& Opportunities}}
\begin{itemize}
    \item \textit{Balancing Privacy and Model Utility:} Privacy-preserving FL systems must carefully balance privacy protection with model performance. It is essential to implement techniques that provide strong privacy guarantees while minimizing the impact on model accuracy.
    \item \textit{Quantifying Privacy Levels:} Establish clear metrics for quantifying privacy levels. All participants should agree on the acceptable privacy thresholds, ensuring that these thresholds align with the collaborative research goals and regulatory requirements. This quantification should be transparent and well-documented to foster trust and compliance among stakeholders.
    \item \textit{Comprehensive Privacy Enforcement:} Privacy protections should be applied uniformly across all components of the FL ecosystem, including clients, central servers, and communication channels. It is also critical to ensure the join or leave of participants does not compromise the federation's privacy promises.
    \item \textit{Empirical and Theoretical Trade-offs:} Address the trade-offs between privacy and model performance requires both theoretical insights and empirical validation. Researchers should focus on understanding how various privacy-preserving techniques impact different aspects of model performance and utility. This includes investigating how privacy budgets, noise levels, and other parameters affect the overall effectiveness of the FL system.
\end{itemize}

\subsection{Fairness and Incentive}\label{sec5-fair-incentive}
In FL for healthcare, research on fairness and incentive mechanisms is relatively underexplored, with only \cite{zhang2024unified} delved into the intricacies of both fairness and incentive in FL for healthcare.
\par
Fairness in FL refers to the equitable distribution of model performance across participants. In healthcare, it often means ensuring that ML models perform consistently across different healthcare providers and demographic groups or patient attributes. These fairness considerations are essential because disparities in model accuracy can lead to unequal treatment outcomes. 
\cite{zhang2024unified} introduced several types of fairness, including horizontal fairness, where different hospitals receive comparable model accuracy, and vertical fairness, which focuses on ensuring that model performance is balanced across different demographic or medical attributes. They also proposed multilevel fairness, which seeks to address both client-level and attribute-level fairness simultaneously, and agnostic distribution fairness, aiming to generalize the model’s fairness to non-participating entities, such as hospitals outside the federation.
\par
Incentive mechanisms are equally important in FL, as healthcare institutions often require motivation to participate in the federation. While regulatory constraints can mandate FL within organizations, voluntary participation in broader FL networks typically relies on clear incentives. For instance, hospitals engaging in FL for tasks like chest radiography classification or COVID-19 detection benefit from access to models that are more accurate than those developed using only local data. A well-designed incentive structure should ensure that participants who contribute more, whether through higher-quality data or computational resources, receive proportionately higher rewards. These rewards could be financial, reputational, or in the form of improved access to infrastructure.

\paragraph{\textbf{Recommendations \& Opportunities}}
\begin{itemize}
    \item \textit{Dynamic Fairness Mechanisms:} Future research should focus on developing dynamic fairness mechanisms that can adapt to changes in data distributions and contributions from different healthcare providers. This would ensure that fairness is maintained even as the data and participants evolve over time.
    \item \textit{Transparent Contribution Metrics:} Establish transparent and robust methods for quantifying each participant's contribution is essential. Accurate measurement of contributions in terms of data volume, quality, and computational resources will facilitate the creation of fair incentive structures.
    \item \textit{Broader Incentive:} Incentive mechanisms should extend beyond financial rewards to include non-monetary incentives, such as reputation enhancement, access to advanced computational resources, and improved patient outcomes. This broader incentive could encourage more diverse participation from healthcare institutions.  
\end{itemize}

\subsection{Evaluation}

\subsubsection{Gaps in Evaluation Metrics}\label{recommend_metrics}
The evaluation of FL models in healthcare heavily focuses on conventional ML metrics, such as accuracy, precision, AUC, sensitivity/recall, specificity, F1-score, Dice score, IoU, HD, and loss value~\citep{dong2018holistic}. 
FL models are typically compared against classical centralized models or localized models, with ablation studies commonly used to isolate the impact of specific modifications. 
Most studies overlooked critical aspects unique to FL, such as communication overhead, resource consumption, privacy, and security concerns, thus failing to capture the complexity of FL systems. 
\par
FL involves frequent communication, which can introduce delays and increase costs. However, only a minority of studies (18\%) included communication efficiency in their evaluation. Metrics such as communication cost, number of communication rounds, and latency are crucial for understanding the effectiveness of FL systems~\citep{hosseini2023proportionally,malik2023dmfl_net}. 
\par
Resource consumption is another crucial factor that has been underexplored, with only 12\% of the reviewed studies measuring computational costs in FL evaluation. Key metrics such as training time~\citep{wang2022privacy}, encryption time~\citep{liu2023bfg}, CPU and memory consumption~\citep{alam2023fedsepsis} are necessary for evaluating the computational efficiency and understanding the feasibility of FL systems~\citep{kalapaaking2022smpc}. 
Without these metrics, it is challenging to comprehensively evaluate the trade-offs between performance and resource requirements, limiting the ability to assess the feasibility and practicality of deploying FL systems in resource-constrained healthcare environments.
\par
Privacy and security evaluations are fundamental for FL in healthcare. Despite this, only 16\% of the reviewed studies assessed these aspects, with methods focusing on vulnerabilities to attacks such as model inversion attacks, and DP guarantees. 
For instance, \cite{jin2023backdoor} evaluated the influence of model inversion attacks on synthetic medical images in FL settings. 
\cite{zhou2024ppml} offered theoretical guarantees for DP to assess the system's privacy resilience.
\cite{hatamizadeh2023gradient} assessed various inversion attacks on medical images to measure and visualize potential data leakage in FL.
\par
Scalability, a critical factor for the widespread adoption of FL in healthcare, has similarly been insufficiently evaluated. Variations in client numbers and patient populations need to be thoroughly assessed to ensure that FL systems can scale across large healthcare networks. Studies by \cite{yan2024cross} and \cite{mullie2024coda} have demonstrated how FL models can handle varying clients and patient populations.
\par
Finally, generalizability and robustness are crucial, especially given the heterogeneity of healthcare data. 
As discussed in Section~\ref{skew}, most included studies focused on narrow aspects of data skew and heterogeneity, limiting the applicability of their findings across diverse healthcare institutions. To fully evaluate the performance of FL in healthcare, models must be tested across varying types of data skew and heterogeneity. Without this, the true potential and limitations of FL in heterogeneous healthcare environments cannot be fully understood.

\begin{table*}[ht!]
    \centering
    \caption{Comprehensive Evaluation Metrics for FL in Healthcare.}
    \resizebox{!}{17mm}{
    \begin{tabular}{c|c}
    \toprule
        Aspect & Recommended Metrics \\
    \midrule
        Performance 
        & Accuracy, AUC, Precision, Sensitivity/Recall, F1-Score, Dice, IoU, HD, Loss \\
    \hline
        Communication Efficiency
        & Communication Cost, Communication Rounds, Latency \\
    \hline
        Resource Consumption
        & Training Time, CPU/Memory Usage, Encryption Time \\
    \hline
        Privacy and Security
        & Attacks listed in Table~\ref{table:fl_risks}, DP guarantees \\
    \hline
        Scalability
        & Client Numbers, Patient Populations \\
    \hline
        Generalizability/Robustness 
        & Data Skew Scenarios listed in Section~\ref{skew} \\
    \bottomrule
    \end{tabular}
    }
    \label{table:fl_metrics}
\end{table*}


\paragraph{\textbf{Recommendations \& Opportunities}}
In summary, there is a pressing need for a more comprehensive evaluation framework for FL in healthcare. This framework should encompass traditional performance metrics, communication efficiency, resource consumption, privacy and security, scalability, and generalizability. Table~\ref{table:fl_metrics} summarizes the key aspects and recommended evaluation metrics that should be considered in future research. By adopting this comprehensive approach, researchers can ensure that FL models are not only effective but also scalable, efficient, and secure for real-world healthcare applications.

\subsubsection{Insufficient Benchmarking}
Numerous FL algorithms have been proposed to address the challenges posed by non-IID data. However, systematic benchmarking of these algorithms is scarce. 
Existing studies employed insufficient data partitioning strategies that failed to capture the diversity of real-world healthcare data distributions. 
Most of them focused on only one or two types of data skew, limiting the scope of analysis and preventing a holistic understanding of algorithm performance under varied conditions.
This limitation extends to other critical aspects of FL evaluation as well. To date, no study has provided a comprehensive evaluation covering performance metrics, communication efficiency, resource consumption, privacy and security, scalability, and generalizability. Such evaluations are essential for a robust understanding of FL's applicability in healthcare.
\par
The absence of comprehensive and universally accepted datasets across various healthcare domains also hinders FL benchmarking. Depending on the research objectives, FL experiments may use datasets that vary significantly in scope and focus, such as medical image classification, segmentation, or reconstruction. Currently, there is no standardized, curated collection of large-scale healthcare datasets across various domains, specifically designed for FL research, which makes it difficult to ensure consistency in benchmarking.
\par
Only one study has reported standardized benchmarking, but it included a limited set of healthcare datasets and failed to integrate key constraints of FL in healthcare, particularly privacy, efficiency, and generalizability~\citep{ogier2022flamby}.

\paragraph{\textbf{Recommendations \& Opportunities}}
\begin{itemize}
    \item \textit{Natural Client Splits and Metrics Definition:} Datasets should incorporate a natural client partition reflecting real-world healthcare scenarios, with clearly defined tasks and evaluation metrics. This will facilitate realistic and meaningful benchmarking.
    \item \textit{Reproducible Train/Test Splits:} Ensure datasets have predefined and documented train/test splits for each client, enabling reproducible experiments and comparisons across different studies.
    \item \textit{Baseline Models for Comparison:} Provide baseline models for each task, including a reference implementation for training on pooled data. This will help researchers compare FL performance against traditional centralized learning approaches.
    \item \textit{Standardized API for FL Algorithms:} Standardize the API for FL algorithms to ensure compatibility with the dataset API, allowing for seamless benchmarking of different FL strategies.
    \item \textit{Framework-Agnostic Algorithm Implementation:} Offer plain Python code for various FL algorithms that is independent of specific FL frameworks, ensuring flexibility and broader accessibility.
    \item \textit{Comprehensive Evaluation:} Cover basic performance metrics, communication efficiency, resource consumption, privacy, security, scalability, and generalizability. Follow the guidelines suggested in Section~\ref{recommend_metrics}.
\end{itemize}

\subsubsection{Lack of Interpretability}
FL models allow decentralized data processing, but their black-box nature makes it difficult to understand how decisions are made. 
The opacity of these models raises concerns about trust and accountability, as medical professionals must be able to explain and justify the decisions made by such systems~\citep{li2022explainable,li2023towards}.
\par
The primary challenge in achieving interpretability in FL stems from the decentralized nature of data. Since each client's data remains private, the global model lacks direct access to local datasets, making it harder to detect biases, noisy features, or irrelevant data points. Additionally, privacy-preserving mechanisms, such as DP, can obscure data details, further complicating efforts to generate meaningful explanations. Moreover, multiple stakeholders are involved in the decision-making process in FL: the central server needs to understand the significance of certain features to ensure reliable global updates, while clients must comprehend how their data contributes to the model's performance. These multi-level interpretability needs, combined with resource constraints (e.g., limited computational power and communication bandwidth), present unique challenges for integrating advanced interpretability techniques in FL.
\par
Interpretable feature selection is an essential component of addressing these interpretability challenges in FL. 
By identifying the most relevant features and filtering out noisy or redundant data, FL models can not only improve performance but also increase transparency in their decision-making processes. In healthcare, this is particularly important, as clinicians need to understand which clinical factors the model considers most relevant. For instance, \cite{soltan2024scalable} utilized SHAP values to analyze the correlation between 20 clinical features and COVID-19 outcomes in FL settings, finding that eosinophil count had the greatest influence on predictions. 
\cite{cassara2022federated} introduced a mutual information-based approach to select relevant features in a decentralized manner, while \cite{zhang2023federated} proposed an unsupervised technique to detect outlier features and group similar ones via hierarchical clustering in FL.
\par
In addition to feature selection, model-specific techniques such as tree-based FL provide further opportunities for interpretability by allowing models to function as ``white-boxes"~\citep{argente2024interpretable,li2024effective}. These methods leverage the model's internal structure to explain its behavior. 
However, such approaches are often highly specific to particular types of data and may not be broadly applicable across different healthcare domains. 
Similarly, while gradient-based explanations and attention mechanisms offer another interpretability route, their effectiveness is sometimes limited due to weak correlations between these methods and the actual decision-making processes of the model~\citep{gong2021ensemble,feng2024robustly}.
\par

\paragraph{\textbf{Recommendations \& Opportunities}}
\begin{itemize}
    \item \textit{Incorporation of Domain Knowledge:} Integrate domain-specific knowledge into FL models to enhance interpretability. In healthcare, leveraging medical knowledge (e.g., known correlations between symptoms and diseases) can help guide feature selection and model design, making it easier to explain model outputs to clinicians.
    \item \textit{Inherit and Expand Explainable Models:} Techniques such as federated decision trees, and rule-based methods could be explored to build models that are transparent by design.
    \item \textit{Client-Side Interpretability Tools:} Create lightweight, client-side tools for interpretability that allow individual clients to better understand how their data contributes to the global model. These tools should be resource-efficient to accommodate the limited computational power and bandwidth of many FL clients, particularly in remote healthcare settings.
    \item \textit{Interpretable Aggregation:} Explore novel aggregation methods that not only combine client updates but also explain why certain updates were prioritized over others. These methods could use techniques such as explainable boosting or weighted aggregation based on feature importance to make the global model more transparent.
    \item \textit{Privacy-Preserving Explainability:} Develop interpretability techniques that align with privacy-preserving requirements in FL, such as DP-aware SHAP values or SMPC for feature importance analysis. These methods should balance transparency with the need to protect sensitive data.
\end{itemize}

\subsubsection{Poor Documentation and Reproducibility}
The reproducibility of FL in healthcare is significantly hindered by several critical issues related to documentation, custom implementations, open-source code availability, and the use of private data.
\par

Firstly, inadequate documentation is a major obstacle. Many included studies lacked crucial details required for reproducing results, such as the methods for data preprocessing, data imputation and augmentation, model initialization, optimization algorithms, and choice of key hyperparameters. This absence of detailed documentation makes it challenging to replicate the reported findings accurately. Additionally, there is often a lack of clarity regarding the data exchanged between clients and the central server. Terms like ``model parameters" and ``model updates" are frequently used without precise definitions, which leads to ambiguity about whether these terms refer to gradients, model weights, or other parameters.
\par
Secondly, the widespread use of custom FL frameworks exacerbates the issue. Many studies chose to develop their own implementations instead of utilizing established, open-source frameworks. Given the complexity of FL systems, custom implementations are more prone to errors and may lack the robustness of well-tested frameworks. This practice can lead to inconsistencies and difficulties in reproducing results.
\par
Thirdly, the availability of open-source code is critically limited. Only 27\% of the reviewed studies made their code publicly available, and none released their trained models. This lack of transparency severely hampers the ability to independently assess and validate model performance, further obstructing reproducibility and impeding future research and application of FL techniques.
\par
Additionally, many studies relied on private data and did not test their methods on publicly available datasets. This practice further complicates reproducibility and fairness in comparisons, as proprietary datasets restrict the ability to conduct fair evaluations and verify results. 


\paragraph{\textbf{Recommendations \& Opportunities}}
\begin{itemize}
    \item \textit{Code \& Model Release:} Prioritize the release of well-documented code and trained models to facilitate independent performance evaluations and advance the field collaboratively. However, to balance transparency with privacy, ensure that shared models incorporate privacy-preserving techniques such as DP or HE to safeguard sensitive information.
    \item \textit{Checklist:} Create a comprehensive FL methodology checklist to improve documentation practices in future studies. The checklist should include guidelines for maintaining transparency while adhering to data privacy standards.
    \item \textit{Pipelines Documentation:} Implement full-stack FL pipelines with documentation to simplify AI development for healthcare institutions, making advanced methods more accessible to users with information technology expertise. Modular and privacy-aware pipeline designs are recommended to reduce the risk of exposing sensitive details.
    \item \textit{Evaluation on Public Datasets:} Encourage the use of public datasets for evaluation to ensure fair comparisons and enhance the generalizability of results. Anonymized or synthetic datasets could also be employed when real-world data cannot be shared openly due to privacy concerns.
    \item \textit{Framework:} Extend existing FL frameworks rather than developing new ones to reduce the risk of errors and support community-driven validation and improvement. Incorporating privacy-preserving modules into these frameworks can address both security and transparency needs effectively.
    \item \textit{Privacy \& Transparency Balance:} Emphasize the importance of balancing open-sourcing efforts with robust privacy measures. Transparency initiatives should focus on sharing aggregated insights, generalizable methodologies, and anonymized results while safeguarding patient data integrity.
\end{itemize}

\section{Conclusions}
In this review, we find that the application of FL to healthcare is still in its relative infancy, with most studies focusing on prediction tasks and often lacking robust demonstrations of clinically significant outcomes.
We delve into the challenges and pitfalls of existing solutions and offer practical guidelines for selecting the most appropriate techniques based on specific application scenarios. 
Additionally, we identify open research challenges that need to be addressed in the near future. We also highlight the importance of establishing standardized methodologies and protocols, as well as promoting the release of open-source code to ensure reproducibility and transparency in FL development in healthcare.
We hope that this review will spark new ideas and inspire numerous possibilities for research and application in healthcare FL.

\section*{Author Contributions}
Method and investigation: M.L., P.X., J.H., and Z.T.;
writing: M.L., P.X., J.H., and Z.T.;
supervision and review: G.Y.
All authors have read and agreed to this manuscript.

\section*{Acknowledgments}
This study was supported in part by the ERC IMI (101005122), the H2020 (952172), the MRC (MC/PC/21013), the Royal Society (IEC\textbackslash NSFC\textbackslash211235), the NVIDIA Academic Hardware Grant Program, the SABER project supported by Boehringer Ingelheim Ltd, NIHR Imperial Biomedical Research Centre (RDA01), Wellcome Leap Dynamic Resilience, UKRI guarantee funding for Horizon Europe MSCA Postdoctoral Fellowships (EP/Z002206/1), and the UKRI Future Leaders Fellowship (MR/V023799/1).

\bibliography{references}
\end{document}